\pdfoutput=1
\documentclass[letterpaper,10pt,conference]{ieeeconf}
\IEEEoverridecommandlockouts
\usepackage{graphicx}
\usepackage{amsmath}
\usepackage{booktabs}
\usepackage{threeparttable}
\usepackage{xcolor}
\usepackage{cite}
\usepackage[hyphens]{url}
\usepackage{tikz}
\usetikzlibrary{arrows.meta}
\title{\LARGE\bfseries A 3D Simulator for Passive SPAD Imaging}
\author{Elia Mergaert$^{1,2,*}$, Rong Zou$^{1,*}$, Roberto Pellerito$^{1}$, Simone Nascivera$^{1}$, and Davide Scaramuzza$^{1}$%
\thanks{$^{*}$These authors contributed equally to this work.}%
\thanks{This work was supported by the European Union's Horizon Europe Research and Innovation Programme under grant agreement No.~101120732 (AUTOASSESS) and the European Research Council (ERC) under grant agreement No.~864042 (AGILEFLIGHT). The authors thank NovoViz for providing the SPAD camera and for their technical support.}%
\thanks{$^{1}$Elia Mergaert, Rong Zou, Roberto Pellerito, Simone Nascivera, and Davide Scaramuzza are with the Robotics and Perception Group, University of Zurich, Switzerland (\texttt{https://rpg.ifi.uzh.ch/})}%
\thanks{$^{2}$Elia Mergaert is with ETH Zurich, Switzerland.}%
}
\begin{document}
\maketitle
\thispagestyle{empty}
\pagestyle{empty}
\begin{abstract}
Passive single-photon avalanche diode (SPAD) cameras provide short-exposure observations for low-light perception, but collecting task annotations for these sensors remains difficult, and annotated datasets are scarce.
We present a 3D simulation pipeline that generates realistic photon detections together with camera-motion and semantic labels, enabling perception models to be trained on synthetic SPAD observations and applied directly to real-world data.
The pipeline couples controllable scene geometry and motion with a sensor-informed observation model that accounts for photon shot noise, dark counts, and spatial crosstalk.
We evaluate the pipeline through calibrated photon-count comparisons and zero-shot sim-to-real transfer across three downstream perception tasks.
We show that classifiers trained only on synthetic SPAD data achieve high accuracy on real low-light recordings at illumination levels less than 2~lux, without real-data fine-tuning.
Similarly, we apply models trained entirely on simulated camera sweeps to real-world camera-rotation estimation and achieve mean absolute rotation errors of less than 1$^\circ$ on real 20$^\circ$ and 30$^\circ$ sweeps at 0.5 and 5.0~lux, when deployed zero-shot.
We additionally train semantic segmentation models on synthetic data with automatically generated dense labels and perform direct inference on real-world low-light recordings; the results further demonstrate the usefulness of synthetic dense supervision for real low-light perception.
Our simulator thus provides a practical source of annotated training data for real SPAD perception.
\end{abstract}
\suppressfloats[t]
\begin{figure}[t]
\centering
\input{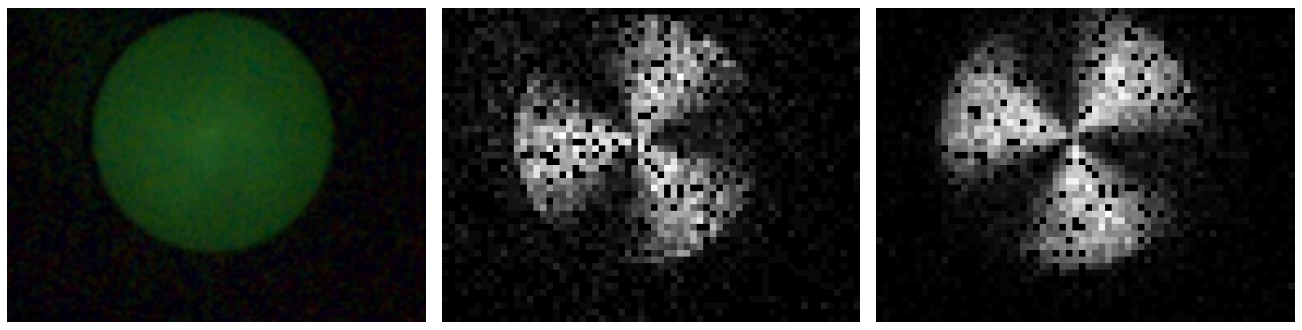}
\caption{Low-light observations of a spinning fan at 1.85~lux. (a) The RGB image has weak contrast and indistinct fan-blade structure. (b) The real SPAD observation preserves recognizable blades within a 1~ms accumulation window. (c) Our simulator generates SPAD observations with similar blade structure and noise appearance.}
\label{fig:teaser}
\end{figure}

\begin{figure*}[t]
    \centering
    \includegraphics[width=1.0\linewidth]{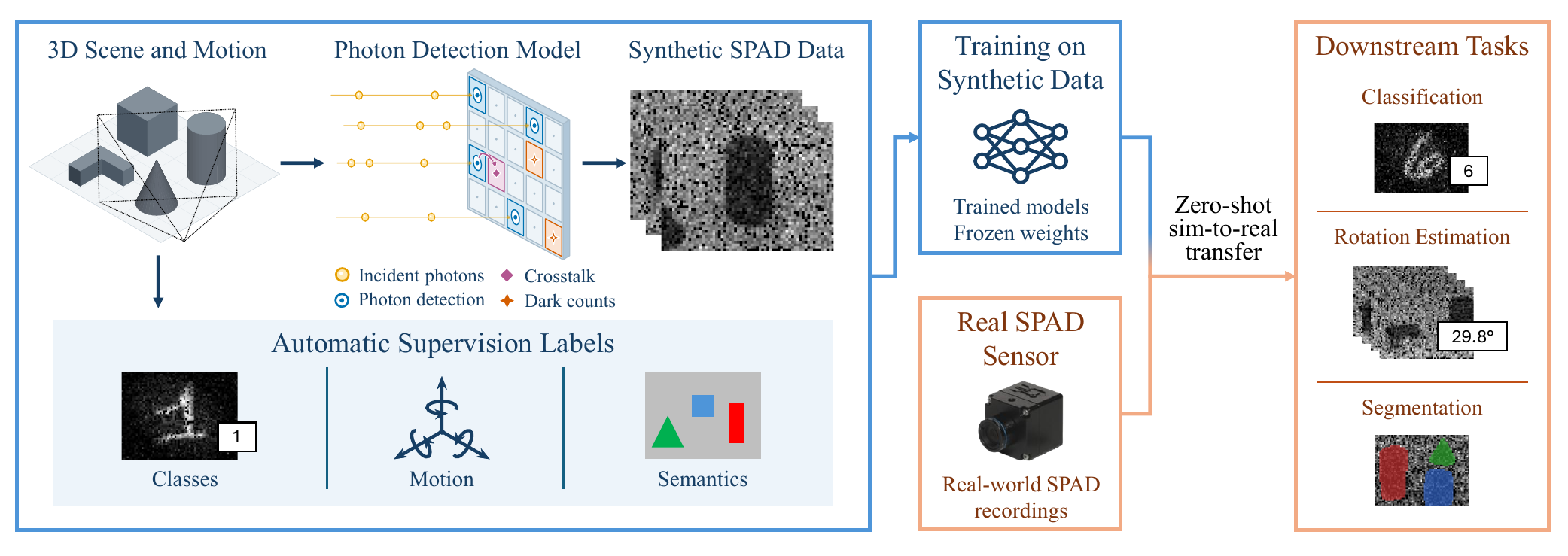}
    \caption{Overview of our 3D SPAD simulator and zero-shot sim-to-real perception pipeline. Controllable 3D scenes and a photon detection model generate synthetic SPAD observations, while scene states provide aligned class, motion, and semantic labels. Models trained exclusively on the synthetic data are applied directly to real SPAD recordings for classification, camera-rotation estimation, and semantic segmentation, with network weights frozen.}
    \label{fig:overview}
\end{figure*}
\section{Introduction}
\label{sec:intro}

Robotic exploration and search-and-rescue missions can require operation in poorly illuminated environments, including underground passages and enclosed spaces~\cite{tranzatto2022cerberus,alismail2017lowlight,chen2024lowlight}.
In these settings, robots must recognize objects, interpret their surroundings, and estimate motion from limited visual information~\cite{mount2016night,crocetti2025comparison,crocetti2025active}.
Low illumination restricts the number of photons available to a camera during each exposure.
More light can be collected by extending the exposure, but this results in blurry observations in dynamic settings~\cite{whyte2012deblurring,ma2020quanta,ma2022deblurnerf,cannici2024mitigating,zou2024retrieval,zou2026eventaided}.
This tradeoff motivates the development of sensing technologies and supporting perception methods that combine sensitivity to weak light with fine temporal sampling.
Passive single-photon avalanche diode (SPAD) cameras offer a promising basis for such perception~\cite{bronzi2016,ingle2019highflux,ma2020quanta,ingle2021,goyal2021photonnet,zhang2025streaming}.

SPADs detect individual photons and support measurements over short time intervals~\cite{bronzi2016,ulku2019swissspad2,morimoto2020megapixel,ingle2021}.
In passive operation, they sense light arriving from the scene without requiring an active illumination source.
Their photosensitivity and temporal resolution have enabled imaging of dynamic scenes under low illumination~\cite{ma2020quanta}.
Fig.~\ref{fig:teaser}(a,b) illustrates this capability by comparing real RGB and SPAD recordings of a spinning fan (5200 rpm) at 1.85~lux. The RGB image shows no discernible blade structure, whereas the SPAD observation preserves recognizable blades within a 1~ms accumulation window.

Translating this strong imaging capability into robotic perception requires algorithms that extract semantic and geometric information from sparse, stochastic photon detections.
Learning-based approaches have demonstrated single-photon image classification and depth estimation, while learned reconstruction has supported object detection and tracking~\cite{goyal2021photonnet,zhang2025streaming}.
These advances highlight the need for training data that connect SPAD observations to the scene properties a perception model must infer.

Obtaining such training data remains a challenge.
The limited availability of annotated SPAD datasets constrains the development of learning-based methods~\cite{suonsivu2026}.
Collecting recordings across object arrangements, viewpoints, motion trajectories, and illumination levels requires repeated acquisition, and each observation must be paired with the appropriate supervision.
Class labels can be assigned to a recorded object, but motion estimation requires synchronized pose measurements, and semantic segmentation requires labels for the visible pixels.
With a moving camera, these targets change with the viewpoint and observation interval, rendering the annotation of physical hardware data time-consuming and difficult to scale.

Simulation offers a way to generate large-scale observations and supervision simultaneously.
In robotic vision, synthetic scenes have enabled models trained in simulation to operate on real images~\cite{tobin2017}, and procedural rendering provides control over scene appearance and camera parameters while automatically producing annotations~\cite{denninger2023}.
For SPAD perception, reference-image simulation has supported digit recognition and demonstrated transfer to real sensor recordings~\cite{suonsivu2024,suonsivu2026}.
Extending synthetic supervision to camera motion and dense scene understanding requires observations to remain aligned with the underlying 3D geometry, camera trajectory, and visible object regions.
A data generator must therefore connect controllable scene evolution with photon measurements and the corresponding task labels.

In this work, we develop a 3D simulator for passive SPAD perception that jointly generates photon observations and task supervision from controllable scenes.
We generate scenes with configurable objects, lighting, and camera trajectories, and convert the rendered signal into photon detections using a photon-detection model that accounts for photon shot noise, dark counts, and spatial crosstalk.
The same scene state supplies object classes, camera ego-motion targets, and pixel-level semantic labels aligned with the generated observations.
This coupling provides automatically annotated SPAD training data for tasks that depend on scene appearance, temporal changes, or dense spatial structure.

We evaluate the simulator through calibrated photon-count comparisons and sim-to-real transfer experiments on image classification, camera-rotation estimation, and semantic segmentation.
For each perception task, network training and model selection use purely synthetic SPAD data, followed by direct evaluation on recordings from a real SPAD camera without fine-tuning.
The experiments demonstrate that our simulator generates realistic data with useful supervision labels for real SPAD perception across multiple downstream tasks.

In summary, our contributions are:
\begin{itemize}
    \item
    \textbf{3D simulation for passive SPAD perception.}
    We develop a SPAD simulator that generates realistic photon observations and aligned ground-truth labels from controllable 3D scenes and camera trajectories, providing high-quality synthetic training data for supervised perception tasks.
    \item
    \textbf{Sim-to-real transfer validation across multiple downstream tasks.}
    We demonstrate the effectiveness of our simulator across image classification, camera-rotation estimation, and semantic segmentation using data captured by a real SPAD camera.
    Models are trained exclusively on simulated data and evaluated on real recordings without fine-tuning.
\end{itemize}
\section{Related Work}
\label{sec:related}

\subsection{Passive SPAD Imaging and Perception}
Passive SPAD cameras have motivated reconstruction methods that recover intensity information from sparse photon detections.
Quanta burst photography aligns and combines rapid binary frames, allowing photon measurements to be accumulated along scene motion~\cite{ma2020quanta}.
Passive inter-photon imaging estimates intensity from intervals between detections, exploiting timing information in the photon stream~\cite{ingle2021}.
Panoramas from Photons iteratively refines motion estimates through temporal aggregation and registration to reconstruct panoramic scenes~\cite{jungerman2023panoramas}.
Together, these methods illustrate the rich information available to vision algorithms through photon statistics and motion.
They motivate simulation pipelines that represent both the sensor response and the changing scene viewed by a moving camera.

Research on single-photon perception further investigates how noisy observations can support inference beyond intensity reconstruction.
Goyal and Gupta introduce photon scale-space and train models for classification and monocular depth estimation with feature consistency across photon levels~\cite{goyal2021photonnet}.
Their approach creates training observations by applying a photon model to existing images, retaining the source dataset's supervision.
Zhang \emph{et al.} maintain a streaming representation over multiple temporal scales, reconstruct images with a neural network, and apply existing detection and tracking models to these reconstructions~\cite{zhang2025streaming}.
These approaches address the design of representations and inference pipelines for photon-limited data.
Our study focuses on generating the observations and labels needed to train these perception models, testing their utility through direct deployment on real recordings.

\subsection{Synthetic Data for Perception}
Synthetic data generation provides a way to obtain supervision while controlling the visual conditions encountered during learning.
Domain randomization varies the appearance and arrangement of simulated scenes to support transfer to real robotic perception~\cite{tobin2017}.
Procedural rendering systems such as BlenderProc2 expose scene geometry, materials, lighting, and camera parameters, and produce annotations alongside rendered images~\cite{denninger2023}.
A shared scene state can therefore determine both an observation and its label, including camera-pose differences or visible object masks.
This is useful for short SPAD observation windows, for which dense annotation or aligned motion measurements can be difficult to obtain from real recordings.
Motion and semantic labels follow the simulated scene state, while the observed contrast also depends on illumination, photon sampling, and sensor noise.
For this setting, scene variation must be coupled with a photon observation model so that training inputs reflect the statistics of the target sensor.

\subsection{SPAD Simulation}
Existing SPAD simulators connect scene appearance to stochastic photon measurements.
Suonsivu \emph{et al.} generate time-resolved SPAD-MNIST from 2D reference images~\cite{suonsivu2024} and model pixel nonuniformities, demonstrating sim-to-real digit classification~\cite{suonsivu2026}.
These image-driven pipelines preserve the labels of their input images but do not generate continuous 3D trajectories or provide dense spatial supervision from an underlying 3D scene.
VisionSIM provides a general framework for scene rendering, ground-truth generation, and passive photon-camera emulation~\cite{visionsim2025}. Its passive SPAD emulator generates binary frames through independent Bernoulli sampling of linearized input intensities, without explicitly modeling dark counts or spatial crosstalk. Our work couples controllable 3D scenes and camera motion with a sensor-informed photon model to generate single-photon observations. The simulator models wavelength-dependent photon detection,
  spatially varying dark counts, and spatial crosstalk, while providing aligned classification, camera-motion, and semantic labels. We assess
our simulator through deployment on real SPAD recordings
across three downstream tasks.

\begin{figure*}[t]
\centering
\begin{minipage}[t]{0.232\textwidth}
\centering\small\textbf{1.85 lux}\\[3pt]
{\footnotesize\makebox[.488\linewidth]{Real}\hfill\makebox[.488\linewidth]{Simulation}}\\[2pt]
\includegraphics[width=.488\linewidth]{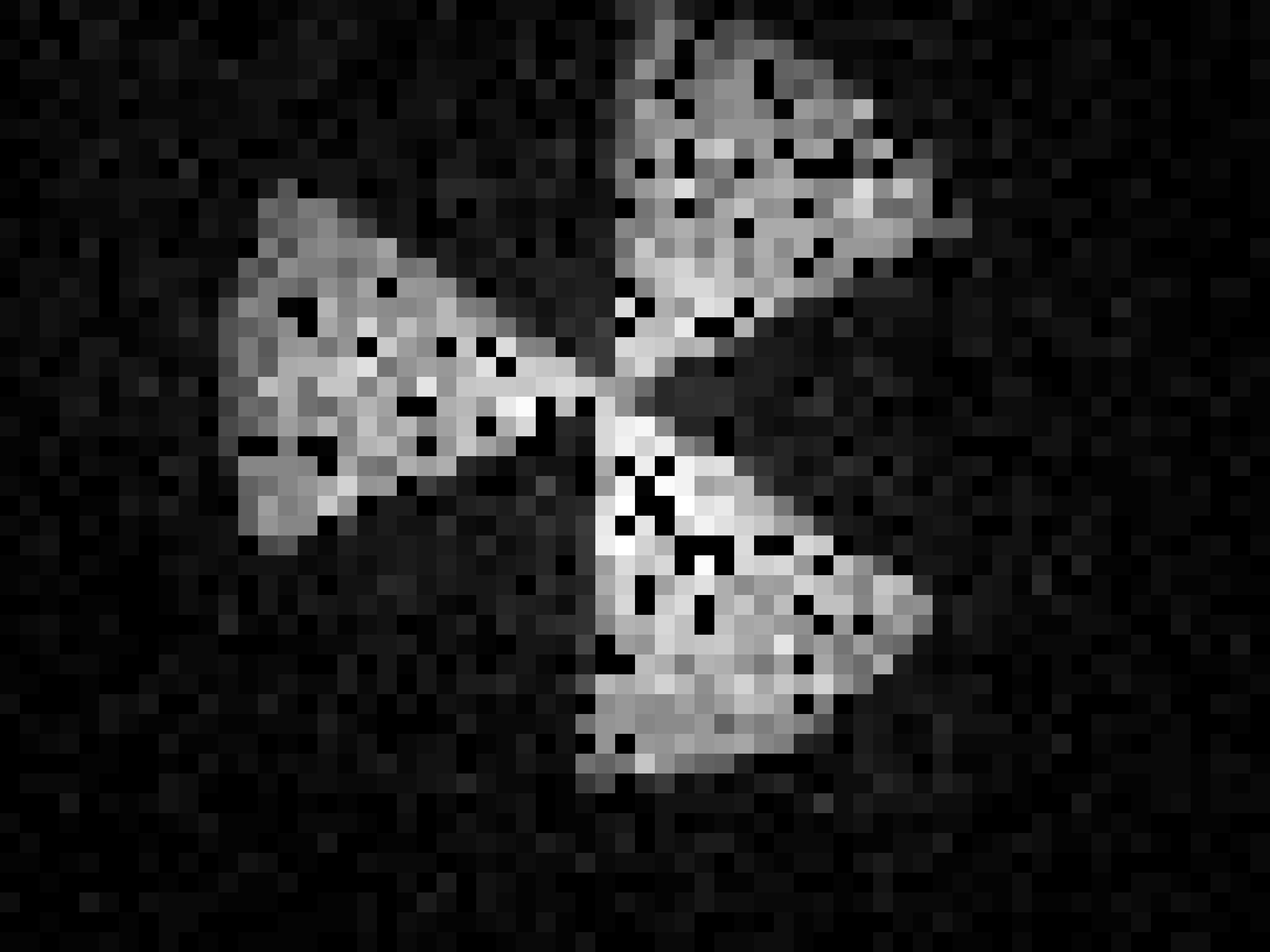}\hfill
\includegraphics[width=.488\linewidth]{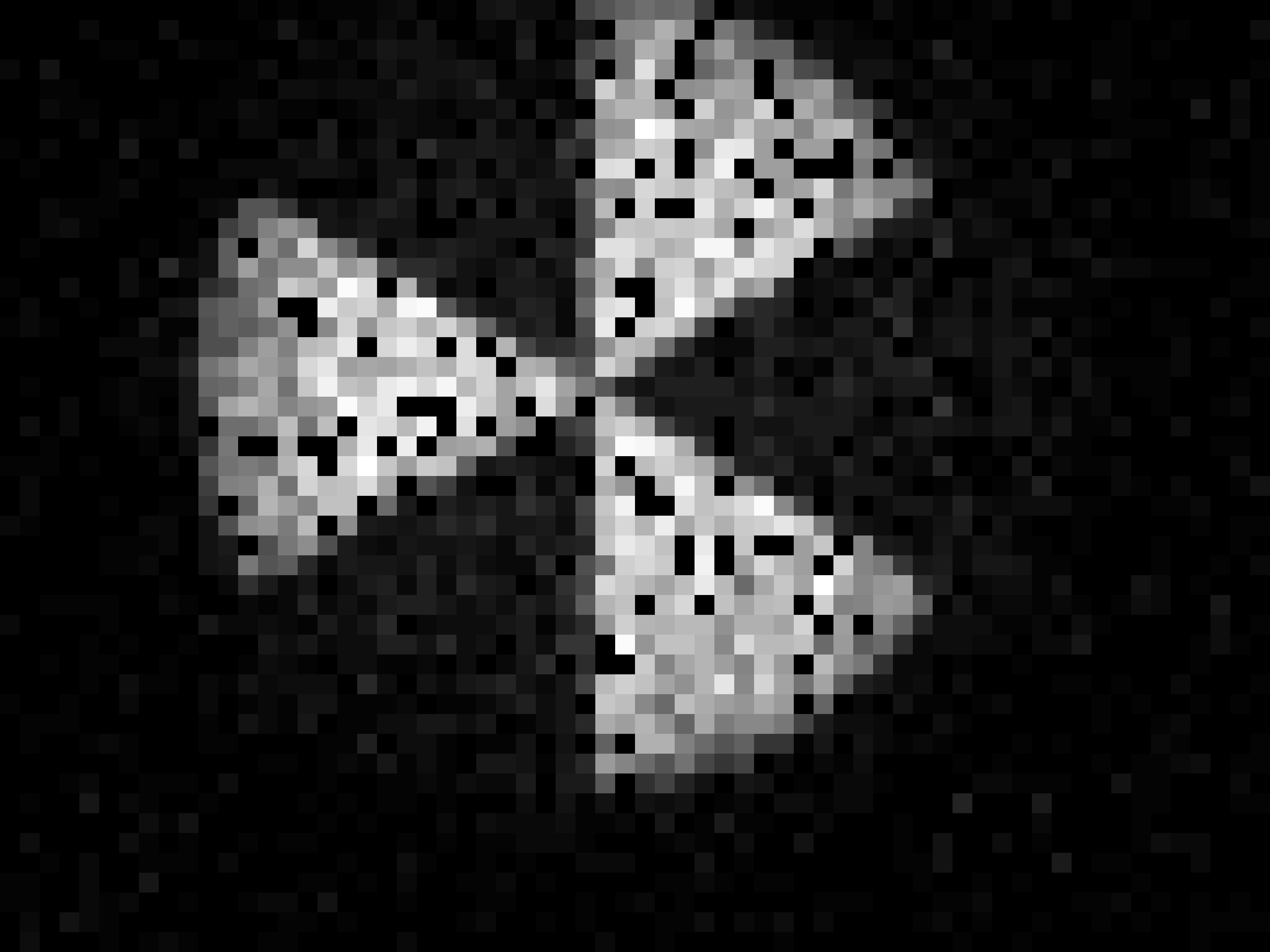}\\[3pt]
\includegraphics{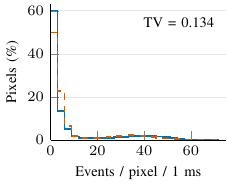}
\end{minipage}\hfill
\begin{minipage}[t]{0.232\textwidth}
\centering\small\textbf{0.40 lux}\\[3pt]
{\footnotesize\makebox[.488\linewidth]{Real}\hfill\makebox[.488\linewidth]{Simulation}}\\[2pt]
\includegraphics[width=.488\linewidth]{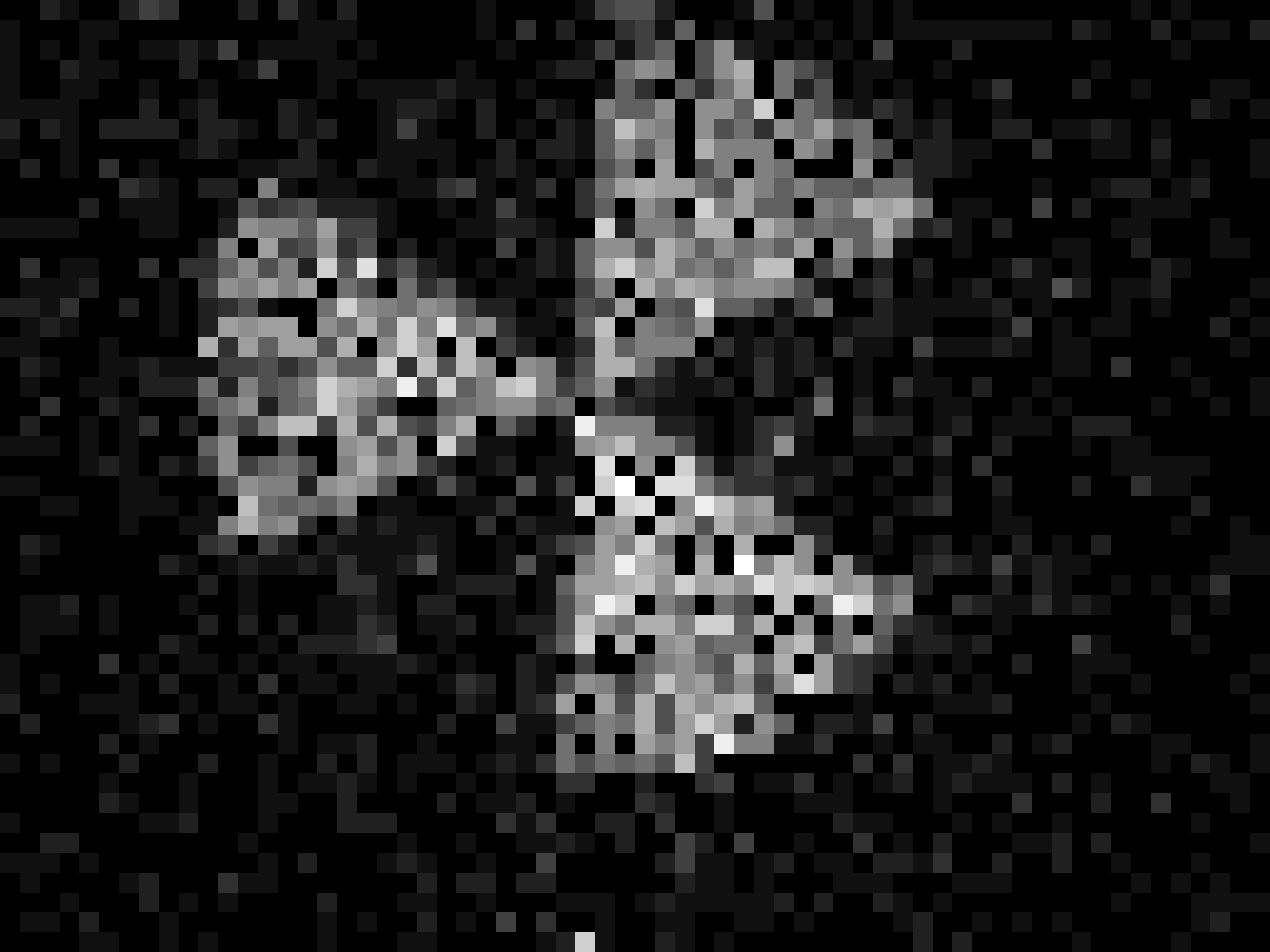}\hfill
\includegraphics[width=.488\linewidth]{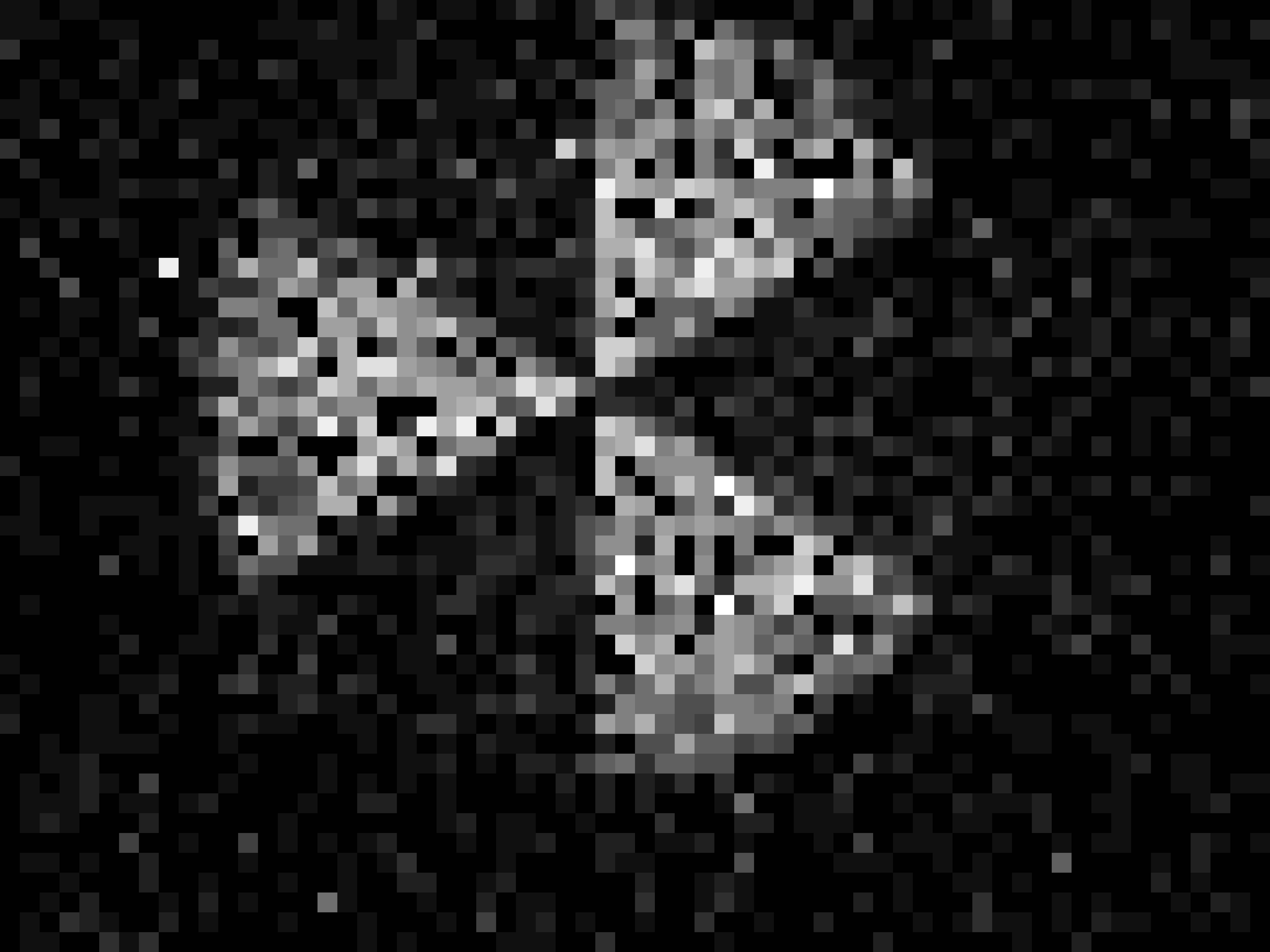}\\[3pt]
\includegraphics{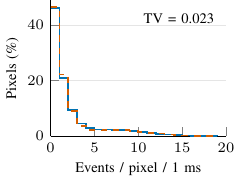}
\end{minipage}\hfill
\begin{minipage}[t]{0.232\textwidth}
\centering\small\textbf{0.15 lux}\\[3pt]
{\footnotesize\makebox[.488\linewidth]{Real}\hfill\makebox[.488\linewidth]{Simulation}}\\[2pt]
\includegraphics[width=.488\linewidth]{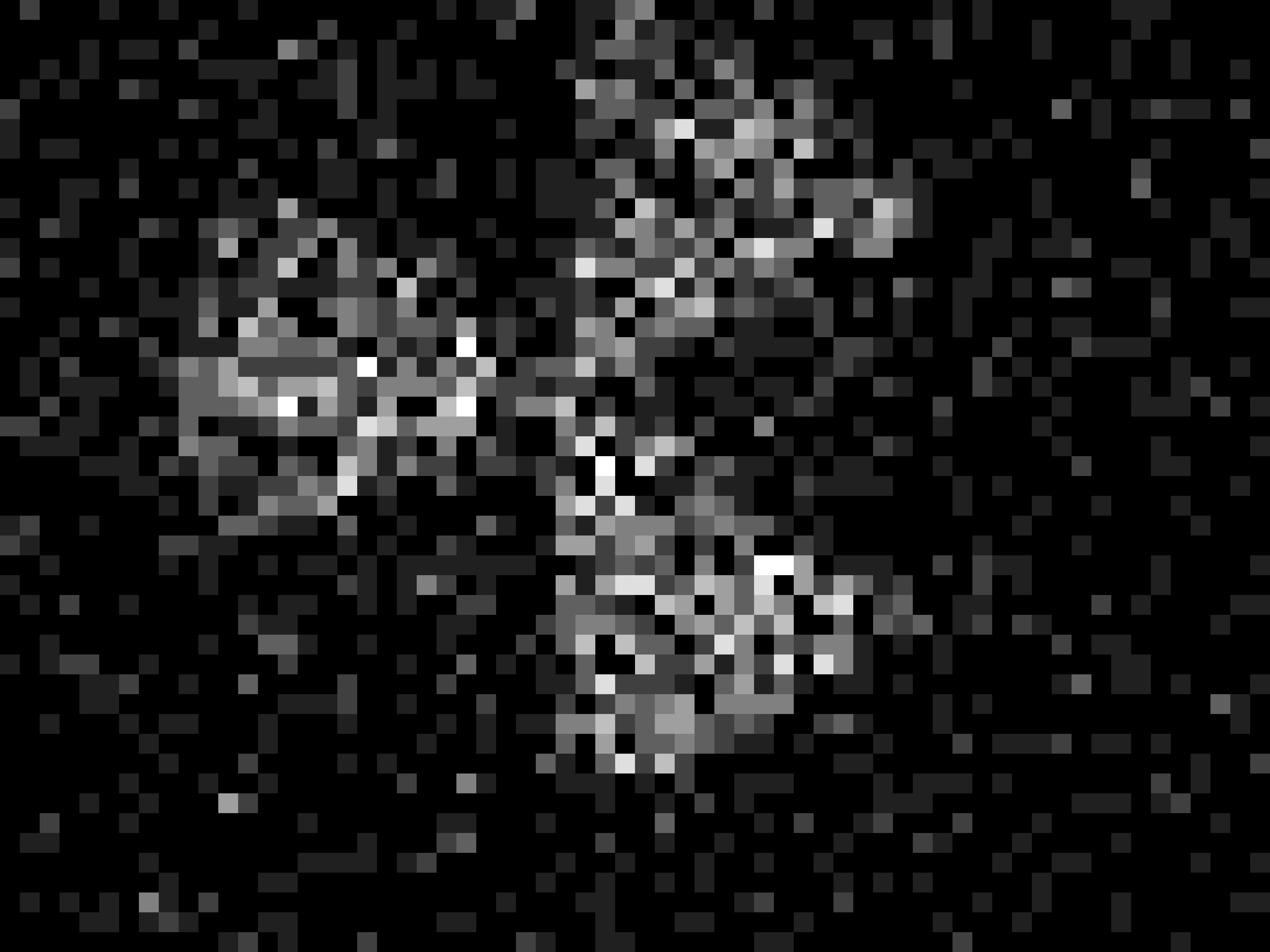}\hfill
\includegraphics[width=.488\linewidth]{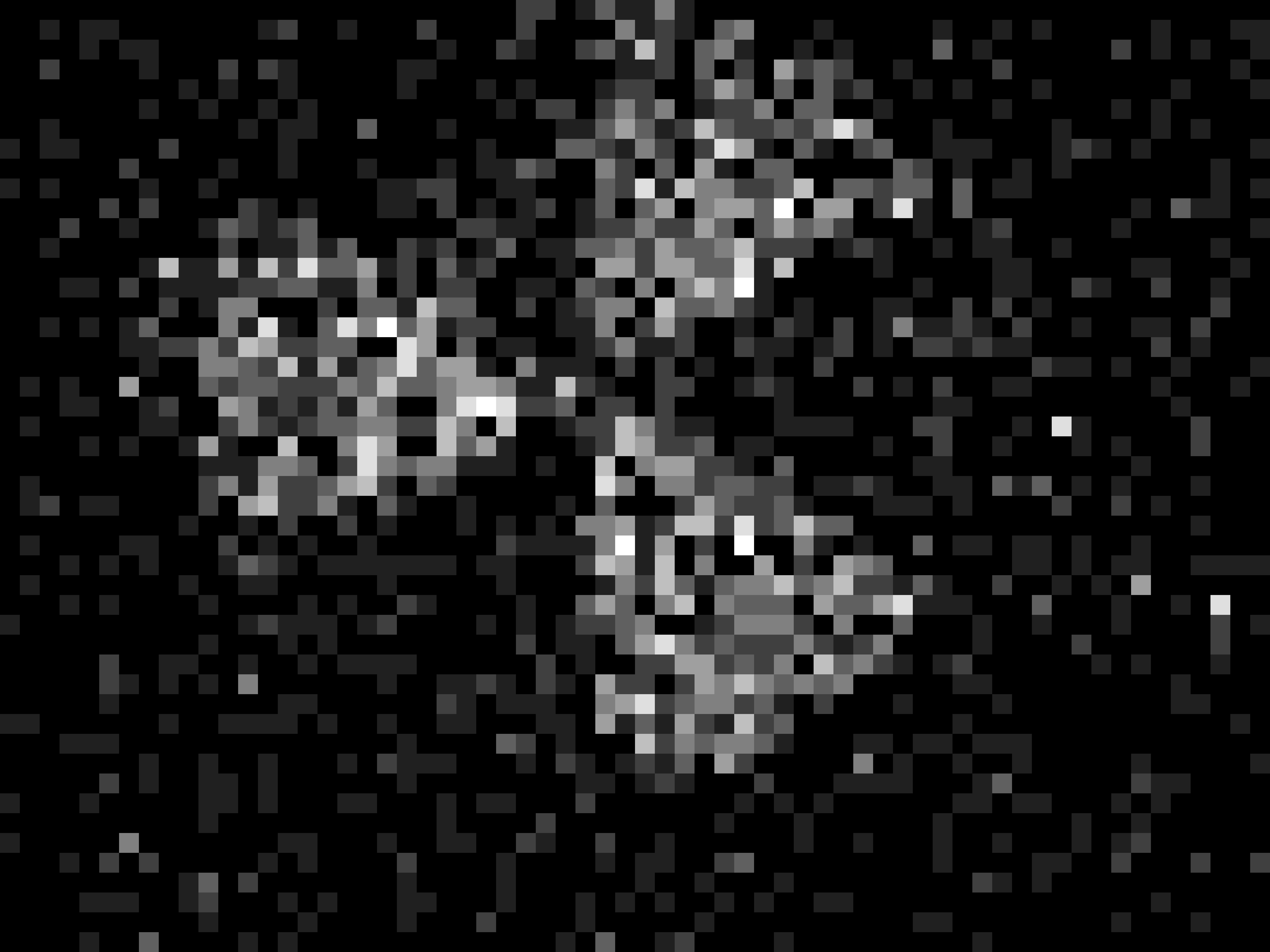}\\[3pt]
\includegraphics{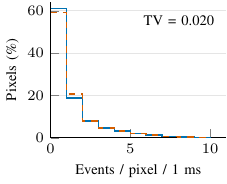}
\end{minipage}\hfill
\begin{minipage}[t]{0.232\textwidth}
\centering\small\textbf{0.05 lux}\\[3pt]
{\footnotesize\makebox[.488\linewidth]{Real}\hfill\makebox[.488\linewidth]{Simulation}}\\[2pt]
\includegraphics[width=.488\linewidth]{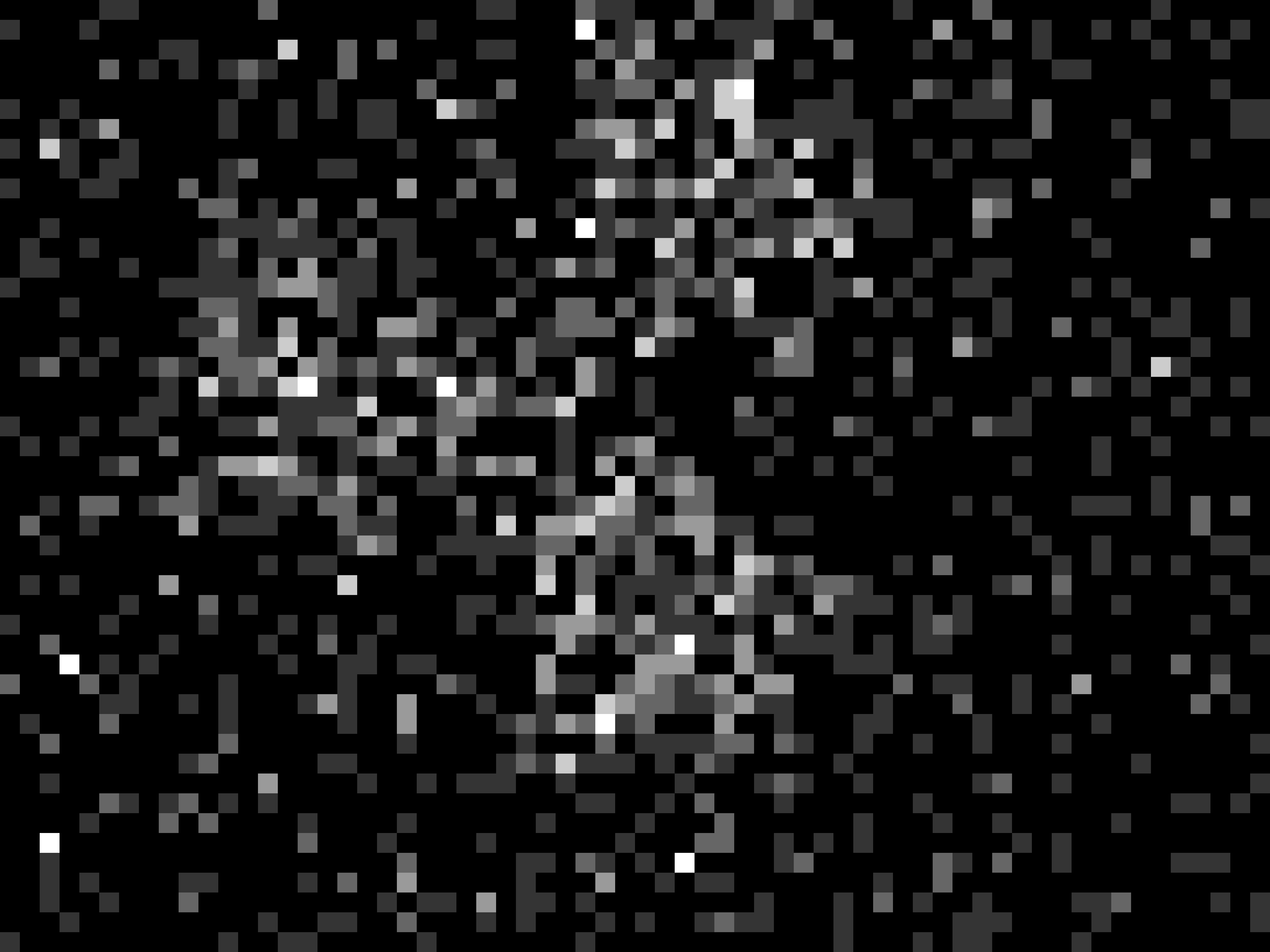}\hfill
\includegraphics[width=.488\linewidth]{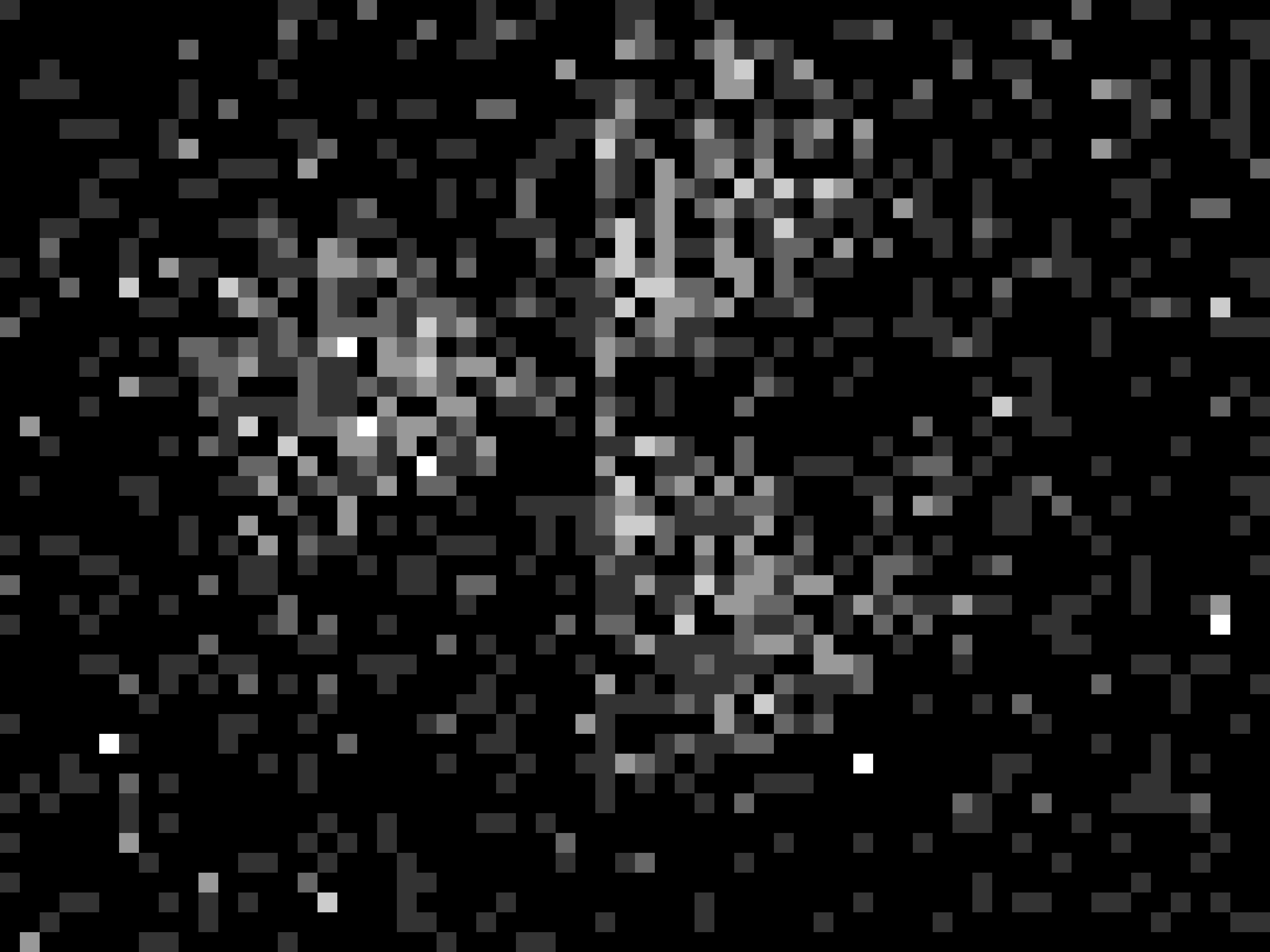}\\[3pt]
\includegraphics{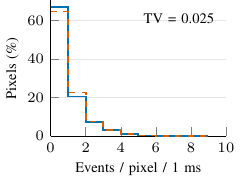}
\end{minipage}
\caption{Visual similarity and photon-count consistency comparisons between real and simulated data across four illumination levels. Top: example 1~ms count frames, normalized for display. Bottom: averaged pixel-count histograms before intensity normalization, shown as a percentage of the $64\times48$ array. \textcolor[HTML]{0072B2}{Blue solid: simulation}; \textcolor[HTML]{D55E00}{orange dashed: real}. Total variation (TV) is computed from the normalized histograms using the original count bins.}
\label{fig:count_validation}
\end{figure*}

\section{3D SPAD Simulator}
\label{sec:method}

Fig.~\ref{fig:overview} outlines our simulator and sim-to-real transfer pipeline.
The simulator maps a time-dependent scene state to photon detections and supervision.
A scene specifies geometry, materials, illumination, and camera pose.
The same state determines both the rendered signal and the ground-truth labels, so labels remain aligned as objects or the camera moves.
We describe the configuration used for the perception experiments below.

\subsection{Scene Control and Supervision}
We define the scene geometry, materials, lighting, and camera trajectories, and render linear RGB responses at the sensor resolution.
For dynamic sequences, we evaluate the scene state at the midpoint of each 1~ms simulation interval and treat the illumination at each pixel as constant within that interval.
For static scenes, we reuse the rendered response while sampling new photon arrivals.

We generate ground-truth labels from the scene state independently of photon sampling.
For classification, we map a digit image to a plane and retain its digit class label.
For motion learning, we compute the rotation between the camera orientations at the first and last observations of a temporal crop.
For segmentation, we assign each visible pixel the class of the object at that location.
Randomized object placement and camera sweeps produce multiple views with corresponding labels, without annotating the noisy observations.

\subsection{Photon Detection Model}
Let $P_\lambda(\mathbf{u},t)$ denote the optical power per unit wavelength incident on pixel $\mathbf{u}$ after optical blur.
Integrating over the sensor's wavelength range $\Lambda$ yields the expected primary detection rate:
\begin{equation}
 r(\mathbf{u},t)=m(\mathbf{u})\left[
 F\int_\Lambda \frac{\eta(\lambda)}{E(\lambda)}
 P_\lambda(\mathbf{u},t)\,\mathrm{d}\lambda+d(\mathbf{u})\right],
 \label{eq:rate}
\end{equation}
where $F$ is the fill factor, $\eta(\lambda)$ is the photon detection probability, $E(\lambda)=hc/\lambda$ is photon energy, $d$ is the dark-count rate, and $m\in\{0,1\}$ masks disabled pixels.
We approximate optical blur by applying a Gaussian point-spread function to the rendered signal before photon sampling.
In our implementation, we approximate the spectral integral with a weighted sum of the linear RGB channels at representative wavelengths of 615, 550, and 465~nm, respectively, using weights $\eta(\lambda_c)/E(\lambda_c)$ for each channel $c$.
The detection probabilities come from the sensor's measured response curve, and a scalar gain maps renderer units to the detection-rate scale.

Dark counts are spatially nonuniform. We sample per-pixel rates from a mean-preserving empirical distribution fitted to sensor dark measurements, with masked hot pixels excluded.
The rate field remains fixed during a simulated sequence, thus capturing the persistent spatial noise pattern while still sampling false detections stochastically over time.
The source dark maps and detection-probability curve are sensor-characterization inputs; they do not provide task labels.

Within each simulation interval of duration $\Delta t$ and midpoint $t_k$, we hold the detection rate constant and sample the number of primary detections as
\begin{equation}
 N(\mathbf{u},k)\sim\operatorname{Poisson}\big(r(\mathbf{u},t_k)\Delta t\big).
 \label{eq:poisson}
\end{equation}
For a pixel with sampled count $N=n$, we draw its timestamps independently within the interval:
\begin{equation}
 \tau_j\mid N=n\overset{\mathrm{i.i.d.}}{\sim}
 \mathcal{U}\!\left(t_k-\frac{\Delta t}{2},\,
                         t_k+\frac{\Delta t}{2}\right),
 \label{eq:timestamps}
\end{equation}
for $j=1,\ldots,n$.

To model spatial crosstalk, for each primary event $e_i=(\mathbf{u}_i,\tau_i)$ we sample
\begin{equation}
 b_i\sim\operatorname{Bernoulli}(p_{\mathrm{ct}}),
 \qquad
 \boldsymbol{\delta}_i\sim\mathcal{U}(\mathcal{D}),
 \label{eq:crosstalk_sampling}
\end{equation}
where $\mathcal{D}=\{(1,0),(-1,0),(0,1),(0,-1)\}$ contains the four neighboring pixel offsets.
If $b_i=1$, we add one simultaneous event
\begin{equation}
 e_i^{\mathrm{ct}}=(\mathbf{u}_i+\boldsymbol{\delta}_i,\tau_i).
 \label{eq:crosstalk_event}
\end{equation}
Events outside the array or at disabled pixels are discarded.
The added events do not trigger further crosstalk.

The pipeline can additionally apply sensor-specific readout arbitration.
This stage is enabled for the static count comparison in Sec.~\ref{sec:counts}.
The learning datasets use the sampled arrival stream before arbitration.
Accordingly, the downstream experiments assess the usefulness of the count-observation model for learning, while the static comparison checks the calibrated sensor-output configuration.

\subsection{Representation and Learning Protocol}
For both simulated and real data, detections are accumulated into nonoverlapping windows of duration $T=1$~ms:
\begin{equation}
 C_k(\mathbf{u})=\sum_i
 \mathbf{1}[\mathbf{u}_i=\mathbf{u}]
 \mathbf{1}[kT\leq \tau_i<(k+1)T].
 \label{eq:counts}
\end{equation}
The learning input is a normalized grayscale count frame, with hot-pixels consistently masked.
Each nonzero frame is independently divided by its own maximum count, quantized to 8~bits, and mapped to $[-1,1]$ for the network.
Zero-count frames remain zero before this final mapping.
The static count comparisons use counts before intensity normalization.

For the experiments, separate models are trained for each reported illumination level.
All networks are fitted on synthetic SPAD inputs, and checkpoints are selected using held-out synthetic validation data.
Real SPAD observations are not used for network training, model selection, or fine-tuning.
The final synthetic evaluation sets are separate from the validation data used during training.
We keep the learning architecture and optimization settings fixed within each task.

\section{Experimental Evaluation}
\label{sec:experiments}

We evaluate the simulator using a NovoViz NV04ASC-HW passive asynchronous camera with a $64\times48$ SPAD array~\cite{novoviz2025}.
We first assess photon-count consistency in a controlled static scene, and subsequently deploy models trained exclusively on synthetic SPAD data to real recordings for image classification, camera-rotation estimation, and semantic segmentation.

The physical camera operates passively without synchronized active illumination, with reference illumination measured directly at the observed scene.
Simulated scenes and camera trajectories are defined in Blender, and renders are generated with Cycles at 4096 samples per pixel.
To match the physical sensor, the nominal simulations apply a fill factor of 0.10 and a crosstalk probability of $p_{\mathrm{ct}}=0.02$.
For all downstream perception tasks, the neural networks are trained using Adam optimization~\cite{kingma2015} for 50 epochs.

\subsection{Calibrated Photon-Count Consistency}
\label{sec:counts}

A static fan with alternating white and black blades is placed 30~cm from the camera in a black enclosure.
We recreate its geometry and viewpoint in the simulator, comparing observations at 1.85, 0.40, 0.15, and 0.05~lux.
Sensor noise is configured using empirical dark-count and hot-pixel characterizations. For this comparison, the dark-count scale and simulated light intensity are set separately for each illumination condition, with the latter aligned to the total recorded photon count.
This experiment therefore measures consistency under these calibrated settings.

Fig.~\ref{fig:count_validation} shows the accumulated count frames and their corresponding pixel-count distributions.
For each condition, let $p_b$ and $q_b$ denote the normalized real and simulated histograms across identical count bins.
We summarize their discrepancy using total variation,
\begin{equation}
 D_{\mathrm{TV}}(p,q)=\tfrac{1}{2}\sum_b|p_b-q_b|.
 \label{eq:tv}
\end{equation}
The total variation distances are 0.134, 0.023, 0.020, and 0.025, corresponding to the decreasing illumination levels.
The agreement is tighter across the three dimmest conditions, while the mismatch is more pronounced at 1.85~lux, especially in the low-count bins. This larger discrepancy may reflect differences in scene-dependent photon flux, whose absolute contribution diminishes as illumination decreases. Overall, these results show that the simulator closely reproduces measured photon-count distributions at different illumination levels.

Qualitatively, the real and simulated observations show similar appearance across different illumination levels: in both domains, the bright blades become less distinguishable from the background as illumination decreases.
To assess whether this visual and statistical agreement supports learning for real perception, we next evaluate sim-to-real transfer in classification, camera-rotation estimation, and semantic segmentation.
Each task uses a fixed dark-count characterization and illumination-specific intensity scales. Within each task and illumination level, these settings remain unchanged across scene instances without scene-specific adjustment.
Models are trained exclusively on synthetic SPAD observations and evaluated on real recordings without fine-tuning.

\begin{table}[t]
\centering
\begin{threeparttable}
\caption{Classification Results}
\label{tab:classification}
\footnotesize
\setlength{\tabcolsep}{3.5pt}
\begin{tabular}{lcccccc}
\toprule
& \multicolumn{2}{c}{1.85 lux} & \multicolumn{2}{c}{0.40 lux} & \multicolumn{2}{c}{0.15 lux}\\
\cmidrule(lr){2-3}\cmidrule(lr){4-5}\cmidrule(lr){6-7}
Training observations\tnote{a} & S & R & S & R & S & R\\
\midrule
Suonsivu et al.~\cite{suonsivu2026} & 61 & 59 & 62 & 58 & 60 & 54\\
VisionSIM~\cite{visionsim2025} & 62 & 60 & 63 & 45 & 64 & 25\\
Ours (Full model) & 63 & \textbf{62} & 61 & \textbf{62} & 61 & \textbf{60}\\
\midrule
Render + Poisson & 64 & 55 & 62 & 44 & 61 & 24\\
No crosstalk & 63 & 58 & 63 & 59 & 61 & 56\\
No dark counts & 61 & 55 & 62 & 40 & 61 & 21\\
Uniform dark counts & 61 & 60 & 63 & 61 & 61 & 47\\
\bottomrule
\end{tabular}
\begin{tablenotes}
\scriptsize
\item[a] Entries are correct predictions out of 64 held-out digits. S: simulated evaluation; R: real evaluation.
\end{tablenotes}
\end{threeparttable}
\end{table}

\subsection{Sim-to-Real Image Classification}
\label{sec:classification}

\paragraph{Task and setup}
We first test sim-to-real transfer for image classification using MNIST~\cite{lecun1998}, whose digit identities can be rendered and physically recorded.
A set of 64 printed digits is recorded at 1.85, 0.40, and 0.15~lux in the enclosure used above, with the simulator mapping the same MNIST digits onto a planar surface in the corresponding scene.
These 64 evaluation identities are excluded from the original 60,000-image training partition, leaving 59,936 synthetic training examples per illumination; renderings of the official 10,000-image test partition serve as synthetic validation data, and both simulated and real evaluations use the same 64 held-out identities.

All models use the same convolutional classifier: three feature stages of 32, 64, and 128 channels, global average pooling, and a two-layer classification head, trained for 50 epochs with cross-entropy, Adam (learning rate $10^{-3}$, weight decay $10^{-4}$), batch size 64, dropout 0.5, and affine augmentation (translation, scale, rotation).
For each observation model and illumination, we select the checkpoint using only synthetic validation accuracy, then evaluate the frozen model on real recordings without fine-tuning.

\paragraph{Comparison of observation models}
We compare our photon observation model with adaptations of two existing SPAD sensor models under identical rendering (scene renderings, geometry, materials, illumination) and learning conditions.
Let $s_i$ denote the expected scene-signal count at active pixel $i$ over 1~ms, after applying the wavelength-dependent detection response, fill factor, and fixed intensity scale, and let $d_i$ denote its expected dark count.

We adapt the sensor model of Suonsivu \emph{et al.}~\cite{suonsivu2026}, whose pipeline starts from reference images rather than rendered 3D scenes, by applying its pixelwise observation model to our rendered inputs. Following their monochromatic formulation, we preserve the total rendered band power, assign it to 555~nm, and apply our camera's detection response at that wavelength to obtain $s_i^{555}$.
Using the same spatial dark-count field as our full model, the adapted count model is $C_i\sim\operatorname{Poisson}(s_i^{555}+d_i)$, independently across pixels.

We also adapt the \emph{VisionSIM SPAD model}~\cite{visionsim2025} to our 1~ms count representation by dividing each interval into $K$ equal subintervals and accumulating their binary outputs; for the static digit scenes, the per-subinterval detection probability is $p_i = 1-\exp(-s_i/K)$, giving $C_i\sim\operatorname{Binomial}\!\left(K,p_i\right)$.
We use $K=8192$ at all illumination levels, chosen using synthetic data to keep binary saturation negligible.

Our \emph{full model} combines the wavelength-dependent detection response, spatially varying dark counts, and neighboring-pixel crosstalk described in Sec.~\ref{sec:method}: primary counts follow $\operatorname{Poisson}(s_i+d_i)$, with additional neighboring-pixel detections generated by the crosstalk model.

All variants share the same active-pixel mask, fixed illumination scales, and classifier preprocessing.
All networks further share the same augmentation, sample training protocol, optimizer, and training budget, with checkpoint selection as above; each frozen model is then evaluated on its own held-out synthetic observations and the same real recordings.

\begin{figure}[t]
\centering
\begin{minipage}[c]{.51\columnwidth}
\centering
\hspace*{15pt}
\begin{tikzpicture}[x=.55cm,y=.55cm,>=Latex,font=\footnotesize]
\node[inner sep=0pt] at (0.25, 0.1) {\includegraphics[width=1.6cm]{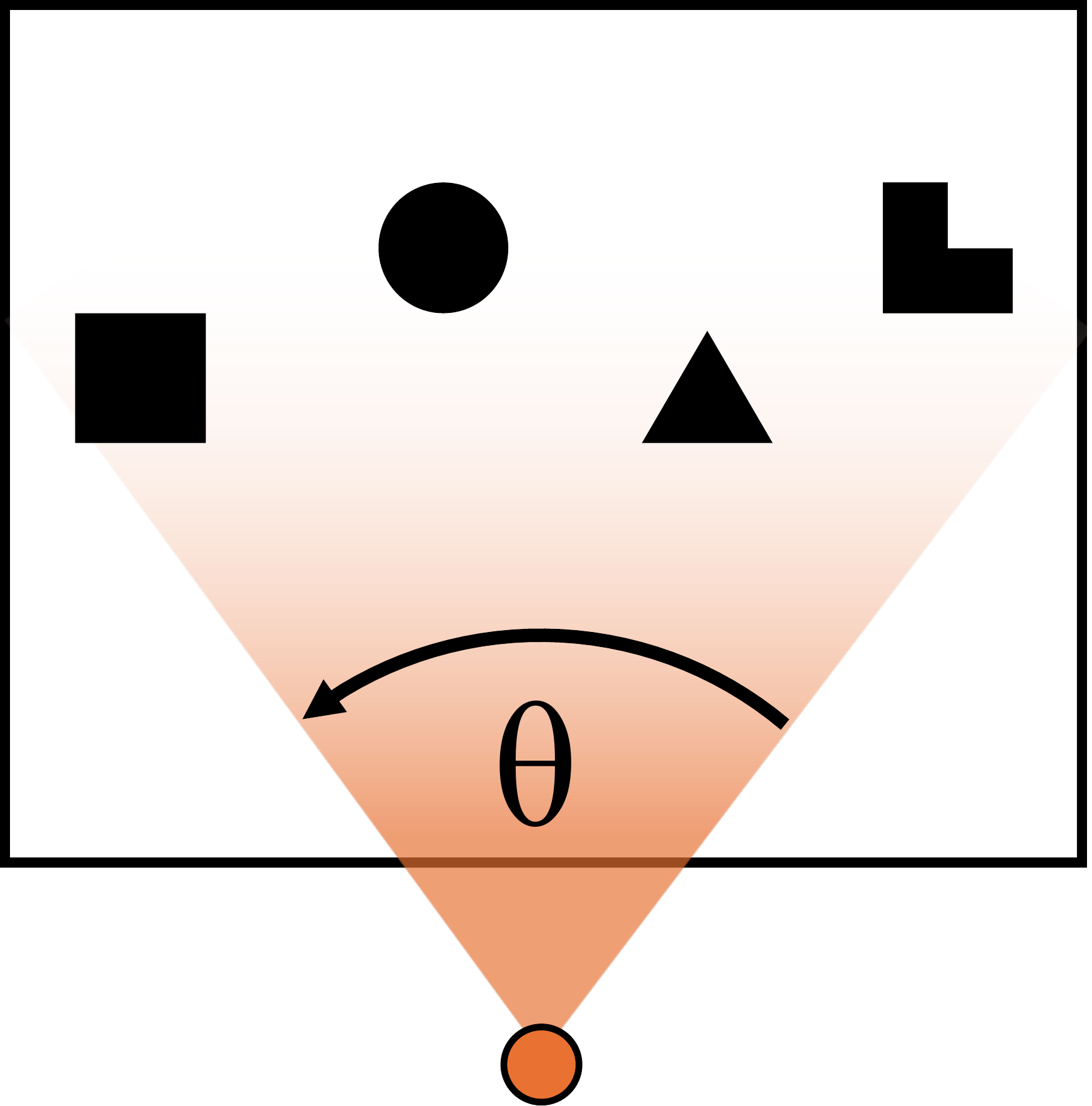}};

\node at (.25,1.95) {Top view};
\node at (.2,-1.65) {Camera rotation};
\end{tikzpicture}
\end{minipage}\hfill
\begin{minipage}[c]{.46\columnwidth}
\centering
\includegraphics[width=\linewidth]{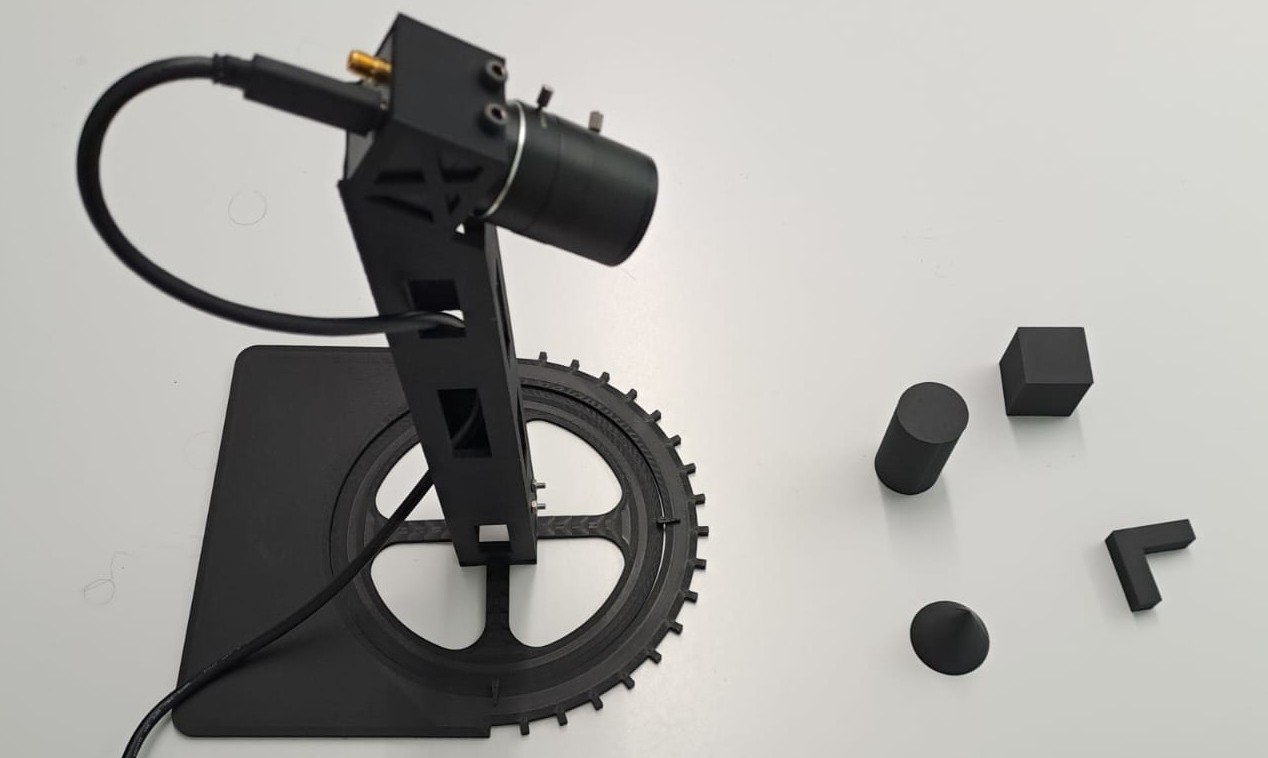}
\end{minipage}\\[4pt]
\makebox[.16\columnwidth][l]{\scriptsize Sim, 0.5 lux}%
\includegraphics[width=.20\columnwidth]{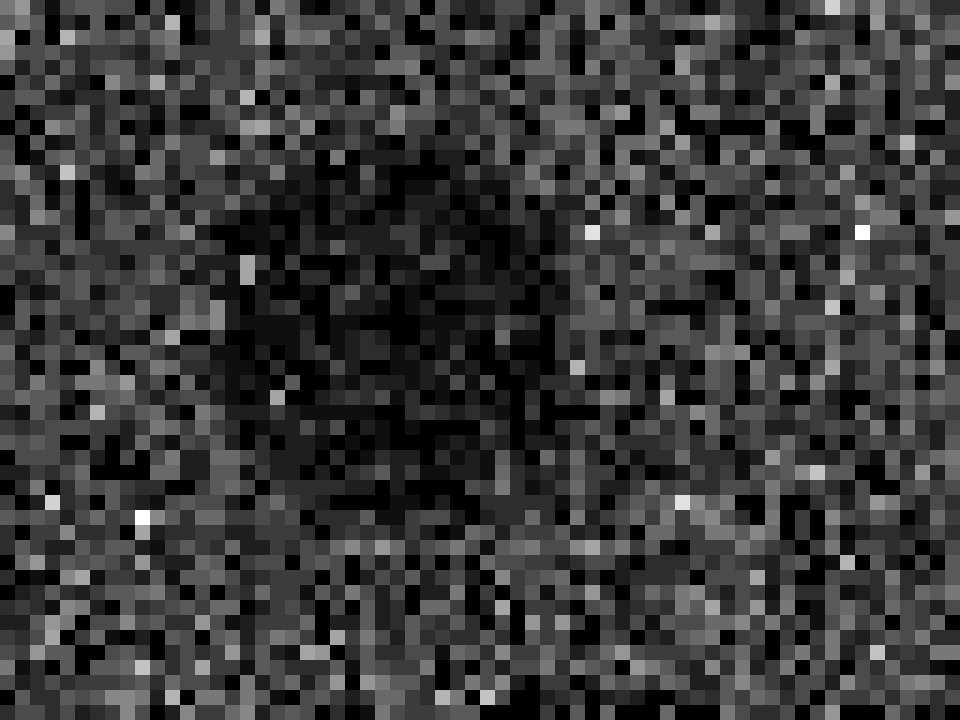}\hfill
\includegraphics[width=.20\columnwidth]{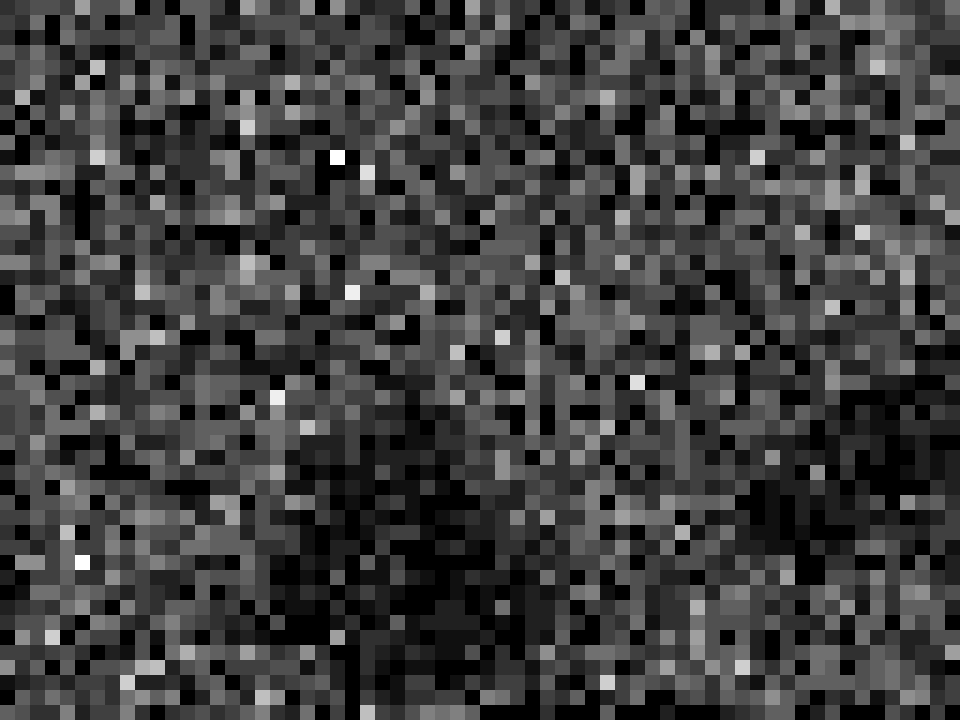}\hfill
\includegraphics[width=.20\columnwidth]{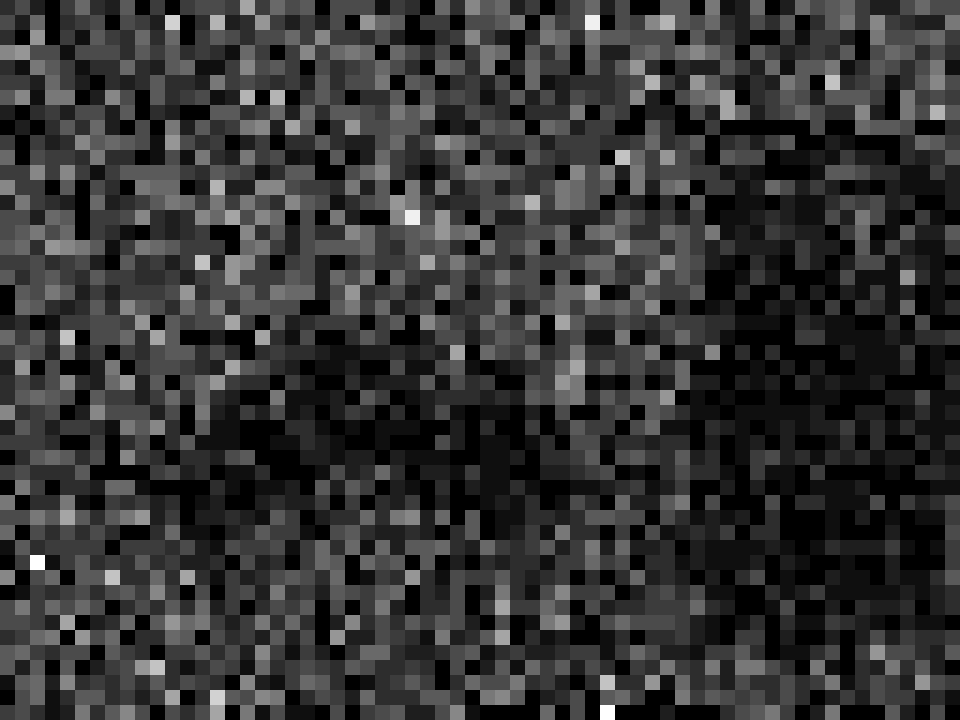}\hfill
\includegraphics[width=.20\columnwidth]{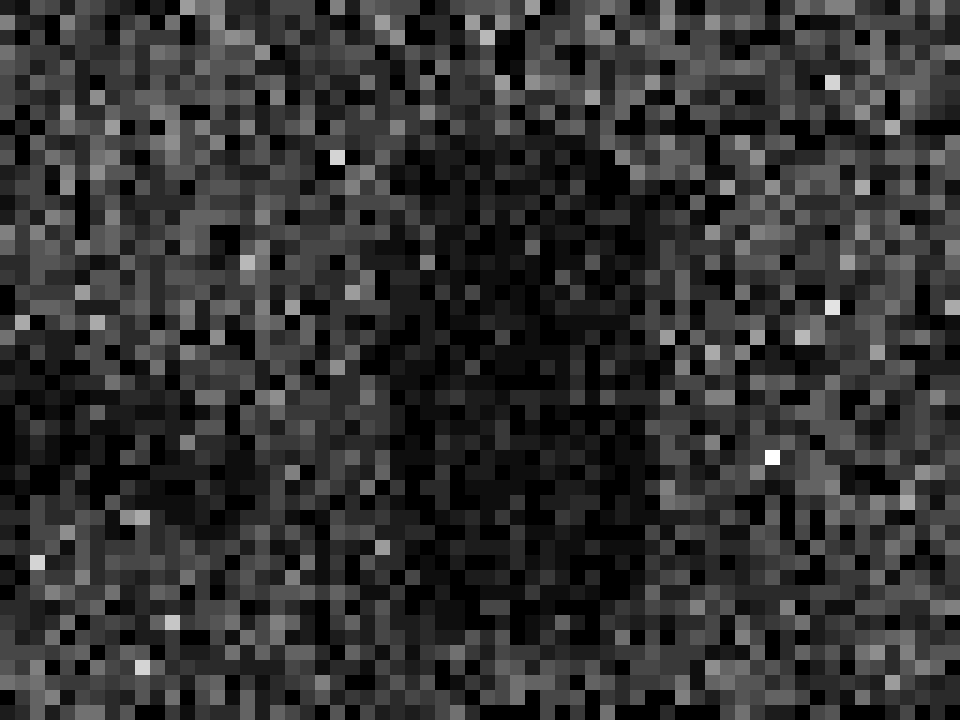}\\[2pt]
\makebox[.16\columnwidth][l]{\scriptsize Real, 0.5 lux}%
\includegraphics[width=.20\columnwidth]{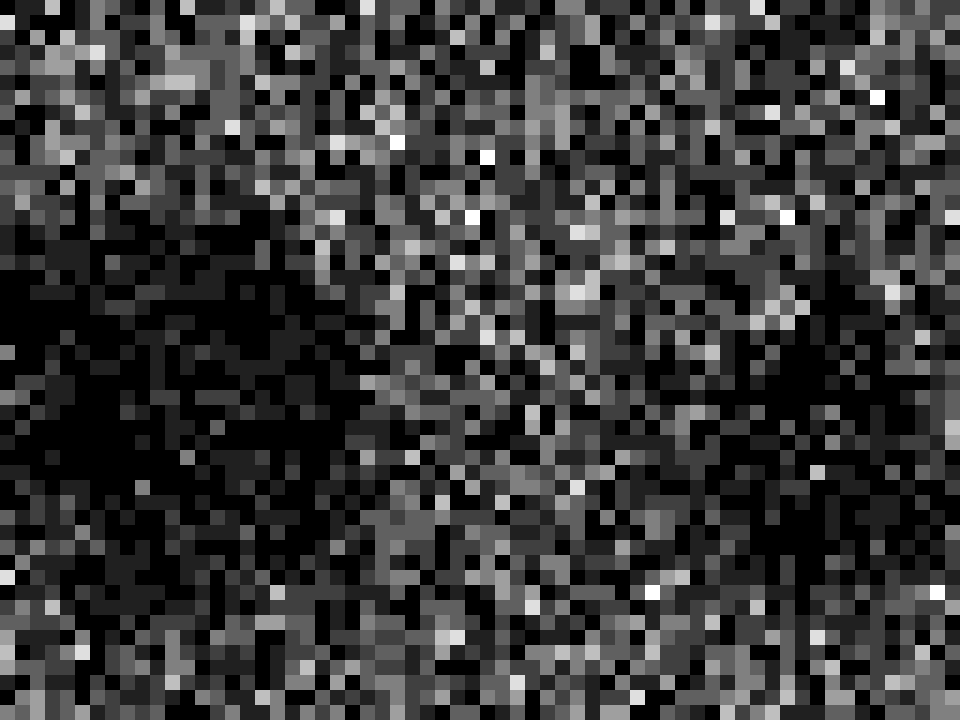}\hfill
\includegraphics[width=.20\columnwidth]{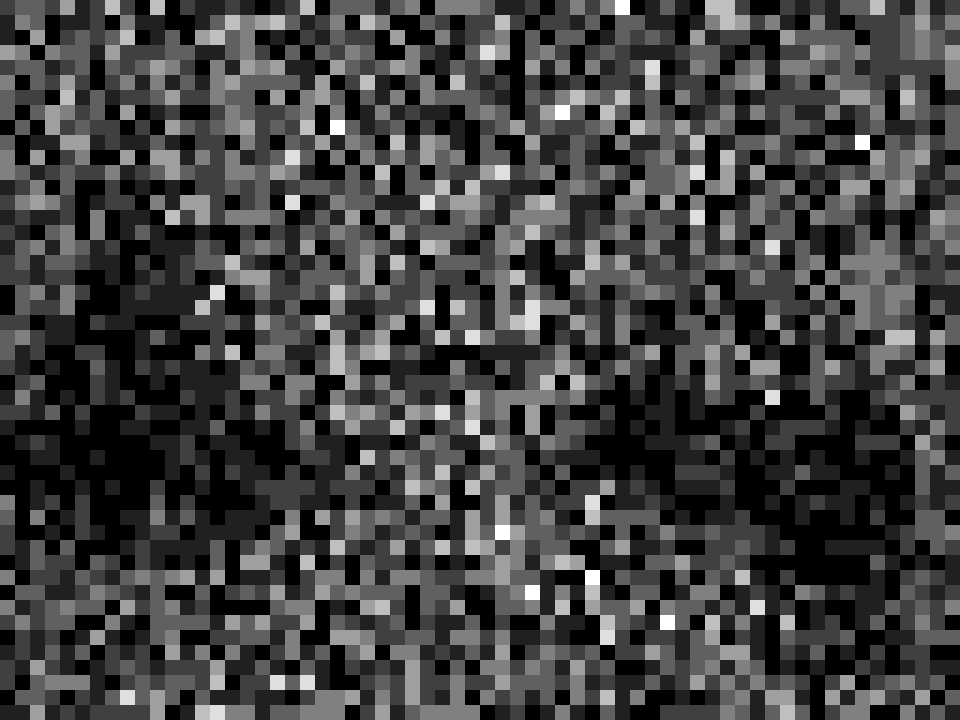}\hfill
\includegraphics[width=.20\columnwidth]{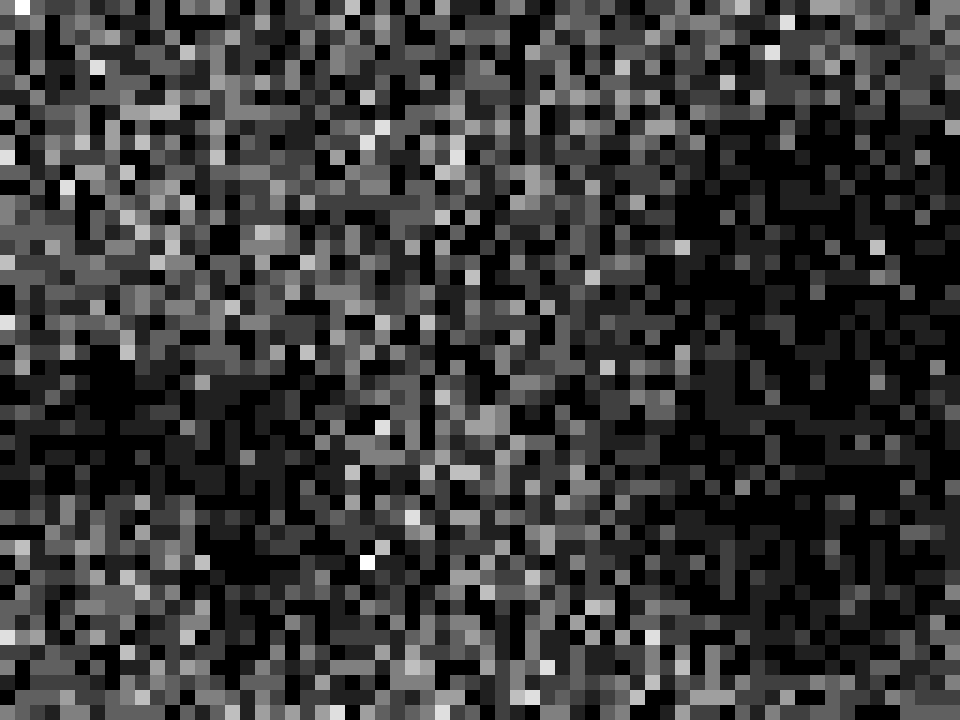}\hfill
\includegraphics[width=.20\columnwidth]{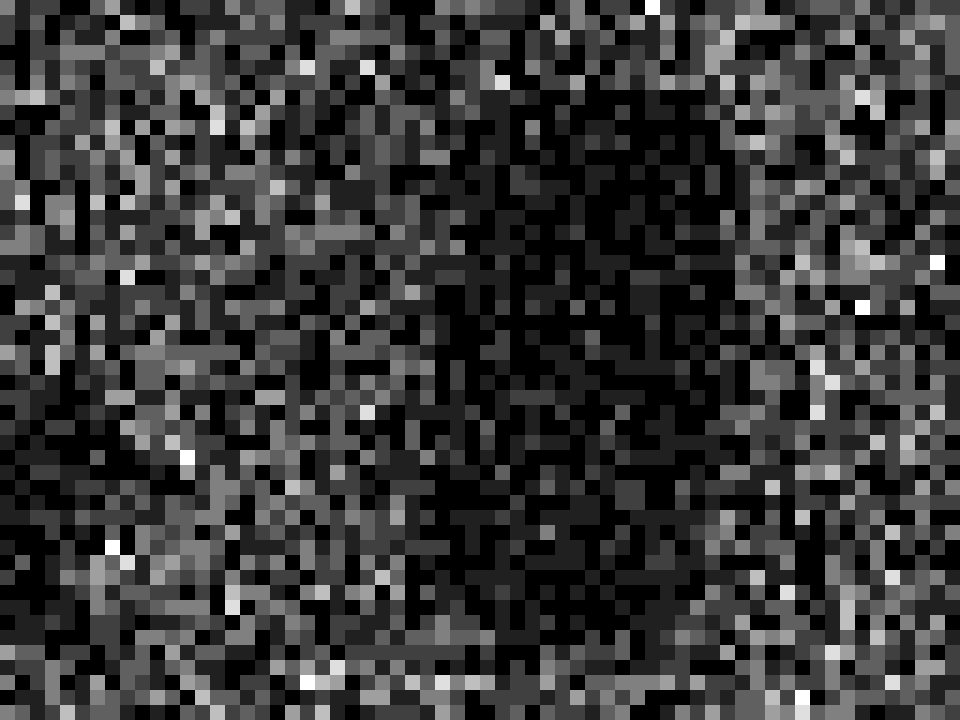}\\[2pt]
\vspace{-5pt}
{\footnotesize\makebox[\columnwidth][r]{Sampled sequence $\longrightarrow$}}\\[2pt]
\caption{Camera-rotation estimation from SPAD sequences. Top: physical tabletop setup and a schematic of the camera sweep. Bottom: example temporal observations at 0.5~lux from a simulated and a real sequence. Photon detections are accumulated over 1~ms; images are normalized for display. The illustrated sequences are separate examples.}
\label{fig:motion_setup}
\end{figure}

\begin{figure*}[t]
\centering
\includegraphics{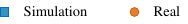}\\[-2pt]
\begin{minipage}[t]{0.485\textwidth}
\includegraphics{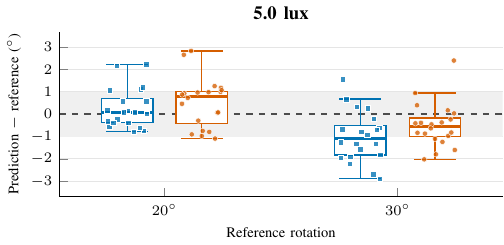}%
\end{minipage}\hfill
\begin{minipage}[t]{0.485\textwidth}
\includegraphics{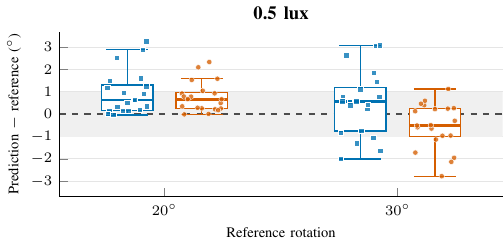}%
\end{minipage}
\caption{Rotation estimation errors for models trained only on simulated SPAD sequences. Each point represents one evaluated sweep; each domain has $n=20$ sweeps per reference angle and illumination. Boxes show the median and interquartile range, with whiskers extending to observations within 1.5 interquartile ranges. The dashed line is zero error; gray shading marks $\pm1^\circ$. All observations, including points beyond the whiskers, are shown.}
\label{fig:rotation_errors}
\end{figure*}

\begin{table}[t]
\centering
\begin{threeparttable}
\caption{Camera-Rotation Estimation Errors}
\label{tab:rotation}
\footnotesize
\setlength{\tabcolsep}{6pt}
\begin{tabular}{cccccc}
\toprule
& & \multicolumn{2}{c}{Simulation} & \multicolumn{2}{c}{Real}\\
\cmidrule(lr){3-4}\cmidrule(lr){5-6}
Lux & Reference\tnote{a} & MAE & RMSE & MAE & RMSE\\
\midrule
5.0 & $20^\circ$ & 0.675 & 0.911 & 0.996 & 1.199\\
5.0 & $30^\circ$ & 1.280 & 1.490 & 0.885 & 1.087\\
0.5 & $20^\circ$ & 0.959 & 1.347 & 0.773 & 1.010\\
0.5 & $30^\circ$ & 1.256 & 1.518 & 0.882 & 1.134\\
\bottomrule
\end{tabular}
\begin{tablenotes}
\scriptsize
\item[a] Errors in degrees ($^\circ$). Each row evaluates 20 simulated and 20 real sweeps. Models are trained and selected using synthetic data.
\end{tablenotes}
\end{threeparttable}
\end{table}

\paragraph{Sensor ablations}
We isolate the roles of dark counts and crosstalk using four ablated variants of the full model, all following the same training and evaluation protocol above.
\emph{Render + Poisson} disables both effects ($d_i=0$, $p_{\mathrm{ct}}=0$), sampling $C_i\sim\operatorname{Poisson}(s_i)$ independently across pixels while retaining the rendered scene appearance and photon shot noise.
\emph{No dark counts} sets $d_i=0$ while retaining crosstalk; \emph{No crosstalk} retains the same spatial dark-count field while setting $p_{\mathrm{ct}}=0$.
\emph{Uniform dark counts} assigns every enabled pixel the full field's mean rate of 959 counts/s, retaining the active-pixel mask and crosstalk, which preserves the expected total dark count before image normalization.

\paragraph{Results}
All observation models achieve at least 93.75\% accuracy on their synthetic evaluations, but their transfer to real recordings differs substantially (Table~\ref{tab:classification}).
The full model reaches 96.88\%, 96.88\%, and 93.75\% real accuracy at 1.85, 0.40, and 0.15~lux, respectively, improving on the adapted Suonsivu model by 4.69, 6.25, and 9.38 percentage points, and on the adapted VisionSIM SPAD model by 3.13, 26.56, and 54.69 percentage points.

The ablations show that modeling sensor noise substantially improves the usefulness of synthetic supervision.
Compared with render + Poisson, the full model improves real accuracy by 10.94, 28.13, and 56.25 percentage points, with the largest gain at the lowest illumination, indicating that rendering scene appearance and adding shot noise alone is insufficient for reliable transfer under low-light conditions.
Dark-count modeling is the major contributor: removing dark counts reduces real accuracy to 85.93\%, 62.50\%, and 32.81\% at the three decreasing illumination levels.
Replacing the spatial dark-count field with a uniform field of the same mean instead causes the largest degradation at 0.15~lux, reducing real accuracy from 93.75\% to 73.44\%, indicating that the spatial structure of the dark-count field matters, particularly under photon-starved conditions.
Disabling crosstalk instead has a comparatively modest effect, reducing real accuracy by at most 6.25 percentage points at any illumination.
Together, these results support sensor-aware observation synthesis, with the strongest gains from dark-count modeling under low illumination.

The following experiments extend the evaluation from static classification to motion and pixel-level prediction, using the temporal and semantic supervision provided by the 3D scenes.

\begin{table}[t]
\centering
\begin{threeparttable}
\caption{Semantic Segmentation Results}
\label{tab:segmentation}
\footnotesize
\setlength{\tabcolsep}{4pt}
\begin{tabular}{lccccc}
\toprule
& & \multicolumn{2}{c}{Simulation} & \multicolumn{2}{c}{Real}\\
\cmidrule(lr){3-4}\cmidrule(lr){5-6}
Setting\tnote{a} & Lux & mIoU & FG mIoU & mIoU & FG mIoU\\
\midrule
Non-occluding & 5.0 & 94.22 & 93.16 & 93.09 & 92.07\\
Non-occluding & 0.5 & 90.99 & 89.38 & 88.48 & 86.79\\
\midrule
Occluding & 5.0 & 52.15 & 40.60 & 54.25 & 43.20\\
\bottomrule
\end{tabular}
\begin{tablenotes}
\scriptsize
\item[a] All scores are percentages. FG: foreground. mIoU averages background and four object classes; foreground mIoU excludes background.
\end{tablenotes}
\end{threeparttable}
\end{table}
\subsection{Synthetic Motion Supervision for Real Camera-Rotation Estimation}
\label{sec:rotation}

\paragraph{Task and reference}
We next evaluate whether motion supervision generated by the simulator transfers to real temporal SPAD observations.
Four black 3D-printed primitives are placed on a white tabletop, and the camera rotates about the vertical axis of its support (Fig.~\ref{fig:motion_setup}).
The task is to estimate the total rotation of a completed sweep.
We evaluate 20 real recordings for each combination of nominal rotation (20$^\circ$ or 30$^\circ$) and illumination (0.5 or 5.0 lux).

The rotation base is clamped in place.
Manually positioned tape markers enter the camera's view outside the intended angular bounds, and frames containing them are removed after recording.
The reference is the nominal sweep angle; the operator estimates an angular uncertainty of at most $1^\circ$ from manual marker placement.
This estimate provides context for interpreting the angular errors reported below.

\begin{figure}[t]
\centering
{\footnotesize
\makebox[.48\columnwidth]{5.0 lux}\hspace{0pt plus 2.3fil}\makebox[.48\columnwidth]{0.5 lux}\\[2pt]
\makebox[.22\columnwidth]{GT}\hspace{0pt plus 1fil}\makebox[.22\columnwidth]{Prediction}\hspace{0pt plus 2.3fil}%
\makebox[.22\columnwidth]{GT}\hspace{0pt plus 1fil}\makebox[.22\columnwidth]{Prediction}}\\[2pt]
\includegraphics[width=.22\columnwidth]{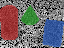}\hspace{0pt plus 1fil}%
\includegraphics[width=.22\columnwidth]{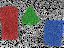}\hspace{0pt plus 2.3fil}%
\includegraphics[width=.22\columnwidth]{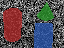}\hspace{0pt plus 1fil}%
\includegraphics[width=.22\columnwidth]{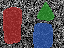}\\[2pt]
\includegraphics[width=.22\columnwidth]{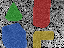}\hspace{0pt plus 1fil}%
\includegraphics[width=.22\columnwidth]{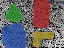}\hspace{0pt plus 2.3fil}%
\includegraphics[width=.22\columnwidth]{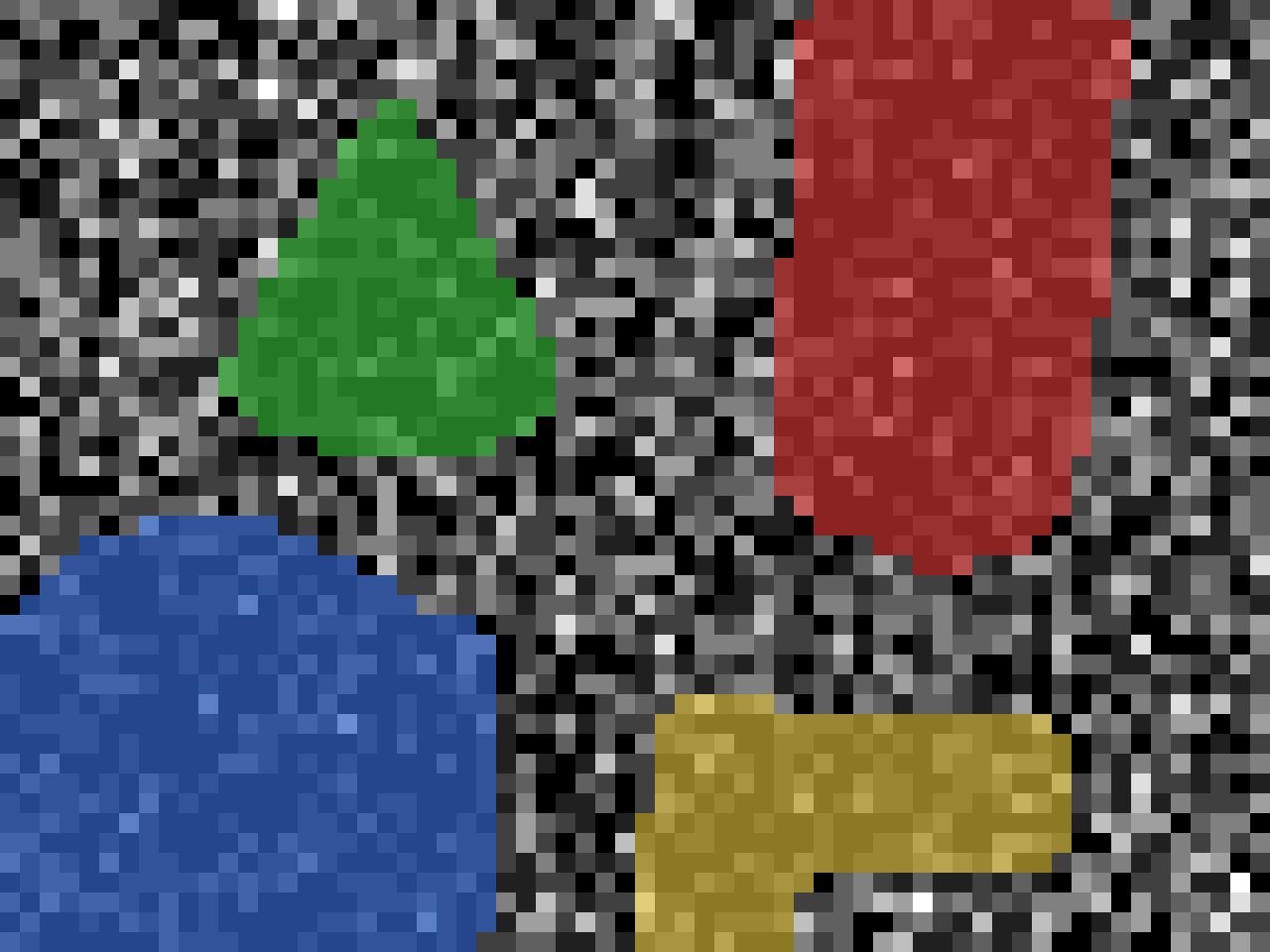}\hspace{0pt plus 1fil}%
\includegraphics[width=.22\columnwidth]{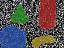}\\[2pt]
\caption{Real SPAD segmentation using models trained entirely on synthetic data. GT shows manually-labeled reference masks, and Prediction shows frozen-model outputs; both are overlaid on the corresponding input. Colors denote cylinder (red), cone (green), cube (blue), and L-shape (yellow).}
\label{fig:segmentation}
\end{figure}

\paragraph{Synthetic supervision}
The virtual scene uses the same primitive geometries with randomized placements.
The camera executes a $30^\circ$ sweep, keeping at least one object visible throughout the trajectory.
For each illumination level, 20 synthetic trajectories are used for training and five separate trajectories for validation.
Ten-frame training crops use strides sampled between 30 and 80 one-millisecond frames, with the difference between camera orientations at the crop endpoints as the regression target.
Varying the crop position and stride provides diverse motion examples.

A compact 3D convolutional regressor uses residual blocks~\cite{he2016} with 32, 64, and 128 channels, global spatiotemporal pooling, and a scalar output head.
Training uses a Huber loss with a 0.05~rad transition, batch size 16, learning rate $3\times10^{-4}$, weight decay $10^{-4}$, and dropout 0.3.
A separate model is trained for each illumination level; checkpoint selection minimizes synthetic validation MAE.
Real recordings are used only for evaluation.

\paragraph{Evaluation}
For each recording, we select 19 approximately uniformly spaced frames, including the first and last observations.
The first ten and last ten form windows $X_1$ and $X_2$ sharing the middle frame, and the total prediction is $\widehat{\theta}=f_\psi(X_1)+f_\psi(X_2)$.
This keeps the network input within the scale of the training crops.
Frame selection uses only the recording length.
We apply the same protocol to 20 held-out synthetic sweeps per condition, disjoint from training and validation.
For errors $e_i=\widehat{\theta}_i-\theta_i$ against nominal sweep references, we report $\mathrm{MAE}=n^{-1}\sum_{i=1}^{n}|e_i|$ and $\mathrm{RMSE}=\sqrt{n^{-1}\sum_{i=1}^{n}e_i^2}$, with $n=20$.
Fig.~\ref{fig:rotation_errors} displays all evaluated predictions as angular errors; Table~\ref{tab:rotation} gives the numerical summary.

\paragraph{Results}
Real-data MAE ranges from 0.773$^\circ$ to 0.996$^\circ$ across the four conditions.
At 0.5~lux, the MAEs are 0.773$^\circ$ and 0.882$^\circ$ for the 20$^\circ$ and 30$^\circ$ sweeps, respectively.
The corresponding real-data RMSE values range from 1.010$^\circ$ to 1.199$^\circ$ across all conditions.
These results demonstrate that motion supervision derived from simulated trajectories supports accurate physical camera-rotation estimates without real-data fine-tuning.

The simulated and real error distributions exhibit comparable scales, although their respective biases differ slightly.
Specifically, the deployed models slightly overestimate 20$^\circ$ sweeps and underestimate 30$^\circ$ sweeps, yielding mean signed errors of approximately $+0.52^\circ$ to $+0.77^\circ$ and $-0.53^\circ$ to $-0.49^\circ$, respectively.
Thus, these results validate the successful transfer of the learned motion relations, even without perfectly identical error distributions across domains.

\subsection{Dense Supervision of Semantic Segmentation}
\label{sec:segmentation}

\paragraph{Task and setup}
The same 3D scene state can simultaneously supply dense labels for each visible pixel.
We evaluate this capability using a U-Net-style model~\cite{ronneberger2015} that distinguishes the four primitive classes and the background.
Synthetic training scenes contain one or two nonoverlapping objects with randomized placement and orientation, while camera sweeps introduce changes in viewpoint.
We generate 250 scenes for training and 50 for checkpoint selection, each sampled at 40 viewpoints, yielding 10,000 training frames and 2,000 validation frames.
Pixel-level object masks are automatically exported alongside the synthetic SPAD frames.
Network training combines weighted cross-entropy and soft Dice losses, using a batch size of 16, a learning rate of $10^{-3}$, and weight decay of $10^{-4}$.
Separate models trained for 5.0 and 0.5~lux are deployed on real data without fine-tuning.

\begin{figure}[t]
\centering
{\footnotesize
\makebox[.32\columnwidth]{Real GT}\hfill
\makebox[.32\columnwidth]{Real prediction}\hfill
\makebox[.32\columnwidth]{Simulated prediction}}\\[2pt]
\includegraphics[width=.32\columnwidth]{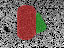}\hfill
\includegraphics[width=.32\columnwidth]{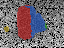}\hfill
\includegraphics[width=.32\columnwidth]{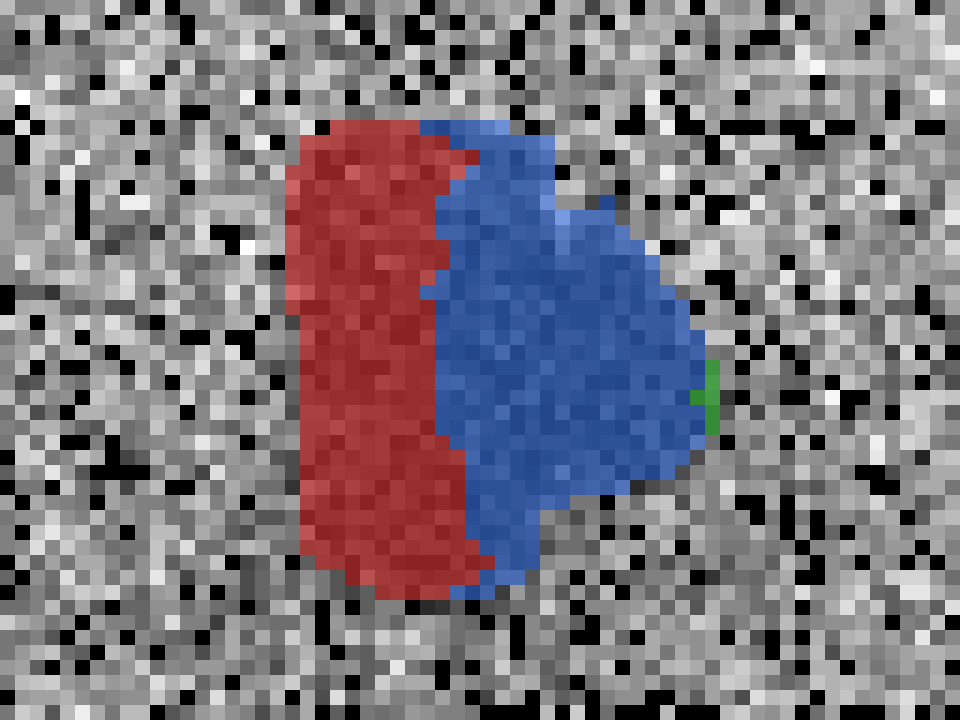}\\[2pt]
\includegraphics[width=.32\columnwidth]{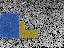}\hfill
\includegraphics[width=.32\columnwidth]{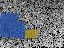}\hfill
\includegraphics[width=.32\columnwidth]{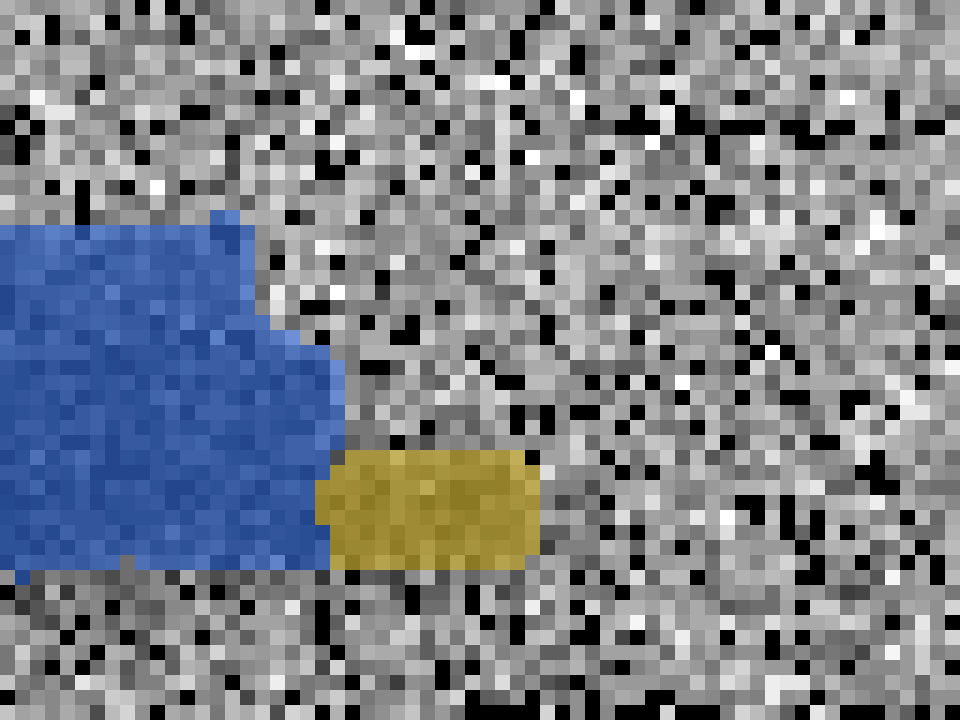}\\[2pt]
\caption{Segmentation predictions on real and simulated SPAD observations at 5.0~lux for object configurations with partial occlusion, which is excluded from the scene layouts used for training. The examples show similar cone-to-cube and L-shape-to-cube confusions. Colors denote cylinder (red), cone (green), cube (blue), and L-shape (yellow).}
\label{fig:segmentation_occlusion}
\end{figure}

\paragraph{Evaluation protocol}
We evaluate 20 real recordings at each illumination, together with two additional recordings of occluding configurations at 5.0~lux.
Reference masks are manually reviewed and corrected on temporally accumulated observations; model inputs remain the original single-exposure SPAD frames.
Each recording contributes one matched input and reference mask, and the model checkpoints remain fixed throughout annotation and evaluation.
For each condition, we pool pixel-level confusion counts across recordings and report mean intersection over union (mIoU) across the five classes.
We also report foreground mIoU over the four object classes to assess object segmentation separately from background prediction. All scores are percentages.
The same frozen models are evaluated on 50 independent synthetic test frames per illumination, with additional synthetic occlusion scenes at 5.0~lux. All synthetic test data are excluded from training and checkpoint selection.

\paragraph{Results}
The models achieve 88.48\% and 93.09\% mIoU on real recordings at 0.5 and 5.0~lux, respectively (Table~\ref{tab:segmentation}).
Foreground mIoU reaches 86.79\% and 92.07\%, showing that the results also hold when background is excluded from the class average.
The corresponding independent synthetic test mIoUs are 90.99\% and 94.22\%; the real-data scores are only 2.51 and 1.13 percentage points lower, respectively.
Although synthetic training scenes contain at most two nonoverlapping objects, the models also segment real scenes containing three or four objects (Fig.~\ref{fig:segmentation}), demonstrating transfer to object arrangements beyond those used for training.
Together, the quantitative results and visual examples demonstrate that automatically generated dense supervision supports semantic segmentation on real low-light SPAD observations without real-data training.

\paragraph{Behavior under object occlusion}
At 5.0~lux, we further evaluate out-of-distribution performance by comparing predictions on real and simulated observations of occluding object configurations absent from the training set (Fig.~\ref{fig:segmentation_occlusion}).
In the cone-to-cylinder configuration, much of the partially visible cone is assigned to the cube class in both domains.
In the L-shape-to-cube configuration, the predicted cube region extends into the upright part of the L-shape in both domains.
IoU metrics for this experiment, given in the last rows of Table~\ref{tab:segmentation}, quantitatively describe this parity, with a difference of at most 3 percentage points.
The similar degradation and class confusions imply that the simulator is able to reproduce the physical scene cues governing the model's behavior.
This supports the use of synthetic scenes to investigate failure modes in real SPAD perception.

\section{Conclusion}
\label{sec:conclusion}

We presented a 3D simulator for passive SPAD perception that generates realistic photon observations and aligned class, camera-motion, and semantic labels from controllable scenes.
By coupling scene geometry and motion with a sensor-informed photon model, the simulator provides automatically annotated data for training perception models.
Calibrated photon-count comparisons show consistency with real sensor recordings.
Experiments across classification, camera-rotation estimation, and semantic segmentation demonstrate that models trained exclusively on synthetic SPAD data can be applied directly to real low-light recordings without fine-tuning.
These results establish our simulator as a practical source of synthetic supervision for developing perception algorithms for real SPAD cameras.
\bibliographystyle{IEEEtran}
\bibliography{references_icra_abrv}
\end{document}